\documentclass[preprint,times,review]{elsarticle}

\usepackage{amssymb}
\usepackage{amssymb}
\usepackage{amsmath}
\usepackage{epsfig}
\usepackage{graphicx}
\usepackage{makecell}
\usepackage{subfigure}
\usepackage{times}
\usepackage{epstopdf}
\usepackage{booktabs}
\usepackage{caption}

\usepackage{hyperref}
\usepackage{float}

\usepackage{verbatim}
\usepackage{algorithm}
\usepackage{algorithmic}
\usepackage{bm}

\usepackage{threeparttable}
\usepackage{multicol}
\usepackage{multirow}
\usepackage{enumerate}

\usepackage{color}

\journal{Pattern Recognition}

\begin{document}
	\captionsetup[figure]{name={Fig.}}
	\begin{frontmatter}

		\title{Vicinal and categorical domain adaptation}

        \author[scut]{Hui Tang}
        \ead{eehuitang@mail.scut.edu.cn}
        \author[scut]{Kui Jia\corref{cor1}}
        \ead{kuijia@scut.edu.cn}
		
		\cortext[cor1]{Corresponding author.}
        \address[scut]{School of Electronic and Information Engineering, South China University of Technology, \\ Guangzhou, Guangdong, China}
		
		\begin{abstract}
		    Unsupervised domain adaptation aims to learn a task classifier that performs well on the unlabeled target domain, by utilizing the labeled source domain. Inspiring results have been acquired by learning domain-invariant deep features via domain-adversarial training. However, its parallel design of task and domain classifiers limits the ability to achieve a finer category-level domain alignment. To promote \emph{categorical domain adaptation (CatDA)}, based on a joint category-domain classifier, we propose novel losses of adversarial training at both domain and category levels. Since the joint classifier can be regarded as a concatenation of individual task classifiers respectively for the two domains, our design principle is to enforce consistency of category predictions between the two task classifiers. Moreover, we propose a concept of vicinal domains whose instances are produced by a convex combination of pairs of instances respectively from the two domains. Intuitively, alignment of the possibly infinite number of vicinal domains enhances that of original domains. We propose novel adversarial losses for \emph{vicinal domain adaptation (VicDA)} based on CatDA, leading to \emph{Vicinal and Categorical Domain Adaptation (ViCatDA)}. We also propose \emph{Target Discriminative Structure Recovery (TDSR)} to recover the intrinsic target discrimination damaged by adversarial feature alignment. We also analyze the principles underlying the ability of our key designs to align the joint distributions. Extensive experiments on several benchmark datasets demonstrate that we achieve the new state of the art.
		\end{abstract}
		\begin{keyword}
			Unsupervised domain adaptation \sep categorical domain adaptation \sep vicinal domain adaptation \sep cross-domain weighting \sep domain augmentation
		\end{keyword}
	\end{frontmatter}
	%\linenumbers
	
	\section{Introduction}
	\label{sec:introduction}
	
    Deep learning of neuron networks \cite{resnet,mnist_lenet} has achieved great success in many machine learning tasks, e.g. image classification \cite{imagenet} %, object detection \cite{focal_loss}, 
    and semantic segmentation \cite{fcn_ss}. These tasks generally assume that data learning and testing models are sampled from the same distribution \cite{tl_survey}. This assumption is easily violated in many practical applications, where data with easy access to labels are often from a domain different from (but related to) that of data with no the access. To well apply a classifier learned on the \emph{source} domain to the \emph{target} one for reducing its labeling cost, domain adaptation \cite{tl_survey} aims to reduce the distribution discrepancy between the two domains. In this work, we focus on the unsupervised setting where %the two domains share the label space and 
    the target domain has no labels.
    
    Domain adaptation theories \cite{da_theory2,da_theory1} state that the expected target error is bounded by the three terms: 1) the sum of the expected source error, 2) a distance metric measuring the distribution discrepancy across domains, and 3) the inconsistency between labeling functions of the two domains. Recent methods focus on minimizing the second 2) or third 3) terms by learning domain-invariant features at the domain or category levels. Among these methods, those based on domain-adversarial training \cite{dann,mada,cdan,bsp,mstn} achieve the current state of the art. They typically adopt a deep network that stacks two parallel classifiers (i.e. the task and domain classifiers) on top of the feature extractor. By adversarial training \cite{gans}, the domain classifier is trained to distinguish features of the source domain from those of the target domain, and the feature extractor is trained to deceive the domain classifier and learn domain-invariant features. 
    
    Despite the progress at reducing domain discrepancy, the parallel design of task and domain classifiers in these methods suggests that the two classifiers, with their corresponding losses, \emph{independently} back-propagate supervision signals, which limits their ability to align the two domains towards the finer category level. In other words, there possibly exists \emph{categorical mismatching} between the aligned source and target domains. Many recent works take steps to mitigate this limitation. For example, MADA \cite{mada} weights the extracted features by corresponding category predictions from the task classifier as inputs of multiple category-wise domain classifiers%, such that each instance is only aligned to several most related categories
    . RCA \cite{rca} matches the same-class labeled source instances and target instances pseudo-labeled by the task classifier, %which correspond to the same category, 
    via a joint domain-category classifier. Based on the joint classifier, SymNet \cite{symnets} enforces the domain- and category-level domain confusions on the target and source domains respectively. However, MADA and RCA only utilize the target-discriminative information from the task classifier and completely ignore that from multiple domain classifiers or the joint classifier, resulting in an unreliable categorical match; besides, the task classifier may be redundant. Especially, SymNet takes no account of the category-level confusion on the target domain and thus a lot of useful information remains to be exploited. 
    
    To overcome these shortcomings, we propose novel adversarial losses at multiple levels on both the source and target domains for \emph{categorical domain adaptation (CatDA)}. Based on the joint classifier that can naturally play the roles of the source and target task classifiers, CatDA applies multi-level adversarial training, where the domain-level one aims to align the whole domains and the category-level one aims to enhance the consistency of category predictions between two task classifiers; category-level adversarial training is technically achieved by a heterogenous, cross-domain weighting design that employs category predictions from the task classifier of one domain to guide the domain-category predictions of the joint classifier on another domain, which can achieve a reliable categorical match. In this work, we also explore a second direction of domain augmentation \cite{gfk,dlow} to push forward domain adaptation. Specifically, we propose to generate a (theoretically) infinite number of augmented domains in the vicinities of the source and target domains, i.e. the \emph{vicinal domains}. Vicinal domains are inspired by \cite{mixup} whose instances are produced by a convex combination of pairs of instances respectively from the source and target domains. Intuitively, the alignment of vicinal domains suggests that of the original domains. We propose novel adversarial losses for \emph{vicinal domain adaptation (VicDA)} based on our adversarial losses for CatDA, leading to our full version method \emph{Vicinal and Categorical Domain Adaptation (ViCatDA)}. 
    Recent works \cite{bsp,tat,da_theory3} tell that adversarial feature alignment could damage the intrinsic discriminative structures of target data. To alleviate it, we also propose \emph{Target Discriminative Structure Recovery (TDSR)} to recover the damaged target discriminative structures, via further fine-tuning the trained ViCatDA model by a semantically anchored spherical k-means clustering algorithm \cite{spherical_k_means}. 
    For deep understanding, we also analyze the working mechanisms of our proposed key designs in principle. Particularly, we explain our cross-domain weighting scheme from an information-theoretic point of view, associated with optimization equilibrium in the two-player game \cite{GenEqu}. 
    In this work, we conduct careful validation studies to verify the efficacy of individual components of ViCatDA and we achieve the new state of the art on several commonly used benchmark datasets. Our main contributions are as follows.
    
    \begin{enumerate}[1)]
    \item We propose novel adversarial losses at multiple levels on both the source and target domains to promote \emph{categorical domain adaptation (CatDA)}. Based on the joint domain-category classifier, the category-level adversarial loss of CatDA improves over the domain-level one by a heterogenous, cross-domain weighting design that enhances the consistency of category predictions between the source and target task classifiers, leading to a reliable categorical match.
    
    \item We propose a concept of vicinal domains and use the vicinal domains to augment the alignment of the original domains. We propose novel adversarial losses for \emph{vicinal domain adaptation (VicDA)} based on our proposed adversarial losses for CatDA, giving rise to the full version of our method termed \emph{Vicinal and Categorical Domain Adaptation (ViCatDA)}.
    
    \item To recover the intrinsic target discrimination damaged by adversarial feature alignment, we propose \emph{Target Discriminative Structure Recovery (TDSR)}, which fine-tunes the trained ViCatDA model by semantically anchored spherical k-means.
    
    \item We also explain the underlying mechanisms of enabling our proposed key designs to reduce the domain discrepancy at a finer category level. Particularly, we explain our proposed cross-domain weighting scheme by connecting it with information theory and optimization equilibrium. 
    
    \item We conduct extensive and careful validation studies to verify the efficacy of individual components of ViCatDA and TDSR. Notably, we achieve the state of the art on several commonly used benchmark datasets.% of Office-31 \cite{office31}, Office-Home \cite{officehome}, VisDA-2017 \cite{visda2017}, and Digits.
	\end{enumerate}

    The rest of this paper is organized as follows. Section \ref{sec:related_works} briefly presents the related works. Section \ref{sec:method} firstly introduces the three closely related methods of MADA \cite{mada}, RCA \cite{rca}, and SymNet \cite{symnets}, and then describes our proposed method in detail. Section \ref{sec:analysis} analyzes our key designs in principle. Section \ref{sec:experiments} shows and discusses the experimental results. Section \ref{sec:conclusion} includes the conclusion and future work.

	\section{Related Works}
	\label{sec:related_works}
	
    %We review the literature in two parts: 1) domain adaptation methods, and 2) domain augmentation ones that can improve the domain adaptation performance.

    \subsection{Domain Adaptation Methods}
    \label{subsec:21}
    
   	Recent unsupervised domain adaptation (UDA) methods can be categorized into the homogeneous and heterogeneous settings \cite{GLG_HeUDA}. %For heterogeneous UDA, Liu \emph{et al.} \cite{GLG_HeUDA} propose a principal angle-based metric to measure both the distance between original heterogeneous domains and the distance between their homogeneous counterparts. 
   	In this work, we focus on the homogeneous UDA setting. 
   	Inspired by domain adaptation theories \cite{da_theory2,da_theory1}, recent UDA methods %aim to %reduce domain discrepancy by 
   	learn domain-invariant deep features at the domain \cite{dann,adda,BeyondSW,cmd,SimNet,gen_to_adapt} or category \cite{mada,cdan,bsp,mstn,symnets,tat,dwt_mec,dan,jan,tpn,gpda,associativeDA,adr,mcd,swd,pfan,dirt_t} %layerWiseCorrection,hla,
   	level. 
   	
   	A popular UDA strategy is to directly minimize the domain discrepancy measured by various metrics. For example, Gretton \emph{et al.} \cite{kernel_two_sample_test} give the theoretical analysis for comparing distributions and present a kernel-based metric of maximum mean discrepancy (MMD)%, which is defined in the reproducing kernel Hilbert space (RKHS) through statistical tests
   	; recently, Liu \emph{et al.} \cite{MMD_D} further advance the development of kernel two-sample test by parameterizing kernels by deep neuron networks. After the seminal work of \cite{kernel_two_sample_test}, many MMD-based UDA methods have emerged, e.g. \cite{BeyondSW,dan,jan,tpn}%,layerWiseCorrection
   	. 
   	For instance, JAN \cite{jan} proposes a joint maximum mean discrepancy (JMMD) criterion, which is reduced to align the joint distributions of multiple domain-specific layers across domains. TPN \cite{tpn} minimizes the distance across prototypes (i.e. class centroids) on data of source, target, and both domains.
   	Other metrics inlcude central moment discrepancy (CMD) \cite{cmd} and association loss \cite{associativeDA}. 
   	
   	Another popular UDA strategy is %to learn more transferable features by 
   	adversarial feature alignment. 
   	Based on the cornerstone \cite{dann}, \cite{adda,SimNet} align whole domains of the source and target%, and thus can only reduce domain discrepancy at the domain level
   	. 
   	CDAN \cite{cdan} utilizes multiplicative interactions between feature representations and category predictions. 
   	MSTN \cite{mstn} and PFAN \cite{pfan} align labeled source centroid and pseudo-labeled target centroid of each shared class. 
   	Some works \cite{adr,mcd,swd} use individual task classifiers for the two domains to detect non-discriminative features and learn discriminative features. 
   	VADA \cite{dirt_t} constrains domain-adversarial training by penalizing cluster assumption violation via entropy minimization. 
   	%Besides global feature alignment, \cite{hla} enforces additional local feature alignment. 
   	%TAT \cite{tat} generates transferable examples to bridge the domain gap. 
   	BSP \cite{bsp} penalizes the largest singular values of feature representations to increase feature discriminability. 
   	GAACN \cite{gaacn} embeds an attention module in GAN to strengthen the discriminator, such that it can distinguish transferable regions among images of the two domains. 
   	CTSN \cite{ctsn} considers the adaptation of tough target samples, by utilizing easy samples and the prediction discrepancy between two individual classifiers. 
   	MADA \cite{mada} and RCA \cite{rca} utilize category predictions from the task classifier to guide the training of category-wise domain classifiers or the joint domain-category classifier on target data, which completely disregard the target-discriminative information from multiple domain classifiers or the joint classifier. 
   	SymNet \cite{symnets} based on domain confusion \cite{simultaneous_transfer} is sub-optimal to achieve category-level domain alignment, since its category-level confusion fully neglects the target-discriminative information. 
   	
   	Other UDA strategies are based on non-adversarial alignment of joint distributions across domains \cite{tat,dwt_mec,gpda}. TAT \cite{tat} freezes the feature extractor of a classification model and trains its task classifier and domain discriminator on corresponding adversarial examples, which fill the domain gap. DWT-MEC \cite{dwt_mec} relies on domain-specific normalization layers to project feature distributions of the two domains to a common spherical distribution. GPDA \cite{gpda} defines a hypothesis space of task classifiers with the Gaussian process and learns prediction consistency via the large-margin posterior separation. Moreover, pseudo-label based methods \cite{atda,ican,dsbn} do self-training \cite{pseudo_label}, which uses the pseudo labels of network prediction as supervision of model training.
   	
   	Differently, our CatDA applies multi-level adversarial training, where the category-level adversarial loss improves over the domain-level one by a heterogenous, cross-domain weighting design that enhances the consistency of category predictions between the source and target task classifiers for both the source and target data, thus promoting the finer category-level domain alignment. 

    \subsection{Domain Augmentation Methods}
    \label{subsec:22}
    
    The previous work \cite{da_for_or} learns a classifier on projected data of the source domain in subspaces whose points are along the geodesic. 
    GFK \cite{gfk} models the domain discrepancy by integrating an infinite number of subspaces along the geodesic flow. 
    DLID \cite{dlid} learns multiple features on augmented domains whose instances are sampled from the source and target domains. 
    Based on mixup \cite{mixup}, recent works \cite{dlow,dm_ada} generate plausibly looking images of intermediate domains by an adversarial loss of GAN types. 
    %Most of them only align the two domains at the domain level.
    
    Differently, our VicDA generates vicinal domains by synthesizing instances along a convex combination path between the original source and target domains, and aligns corresponding vicinal domains of the source and target, which can be naturally combined with CatDA to enhance its alignment accuracy. 
	
	\section{Method}
	\label{sec:method}
	
	Given $\{ (\mathbf{x}_i^s, y_i^s) \}_{i=1}^{n_s}$ of labeled instances sampled from the source domain ${\cal{D}}_s$, and $\{ \mathbf{x}_j^t \}_{j=1}^{n_t}$ of unlabeled instances sampled from the target domain ${\cal{D}}_t$, unsupervised domain adaptation aims to learn a feature extractor $G(\cdot)$ and a task classifier $C(\cdot)$ such that the expected target error $\mathbb{E}_{(\mathbf{x}^t, y^t) \sim {\cal{D}}_t}\left[ {\cal{L}}_{\mathrm{cls}}(C(G(\mathbf{x}^t)), y^t) \right]$ is low for a specified classification loss ${\cal{L}}_{\mathrm{cls}}(\cdot)$. Suppose the classification task has $K$ categories, and accordingly $y^s, y^t \in \{1, 2, \cdots, K\}$. Since the two domains by assumption follow different distributions, the main challenge is to minimize the domain discrepancy such that labeling on the source domain can be transferred to the target domain to minimize its error.
	
	State-of-the-art methods are based on domain-adversarial training \cite{dann,bsp}. These methods are usually based on a deep network comprising convolutional (conv) and fully-connected (FC) layers, where the lower conv layers are used as the feature extractor $G(\cdot)$, upper FC layers are used as the task classifier $C(\cdot)$, and a domain classifier $D(\cdot)$ of FC layers is also used on top of $G(\cdot)$, which is in parallel with $C(\cdot)$% and designed for domain-adversarial training
	. The adversarial signal of domain discrimination provided by $D(\cdot)$ aims to make features learned at $G(\cdot)$ become domain-invariant, such that they are ready for use by $C(\cdot)$ for classification of data on the target domain. However, the parallel design of $C(\cdot)$ and $D(\cdot)$ suggests that they \emph{independently} back-propagate supervision signals; even though domain-adversarial training of $D(\cdot)$ would align at $G(\cdot)$ the source and target features \emph{as a whole domain}, the alignment is not expected to go finer to the category level, i.e. there possibly exists \emph{categorical mismatching} between the aligned source and target domains. Many of recent efforts are devoted to alleviating this issue, e.g. \cite{mada,rca,symnets}. 
	
	In this section, we first briefly introduce the three closely related works. Then, we describe our proposed method in detail. 

	\subsection{Brief Introduction of Closely Related Works}
	\label{subsec:31}
	
\begin{figure}
    \centering
    \subfigure[MADA]{
        \label{fig:mada}
        \includegraphics[width=0.85\linewidth]{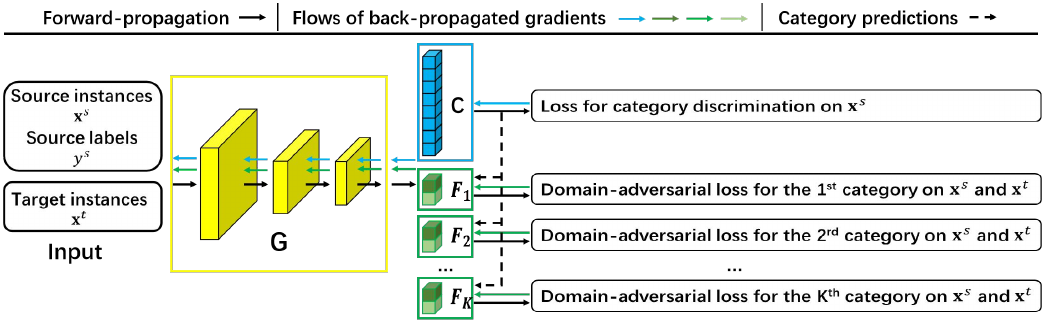}
    }

    \centering
    \subfigure[RCA]{
        \label{fig:rca}
        \includegraphics[width=0.85\linewidth]{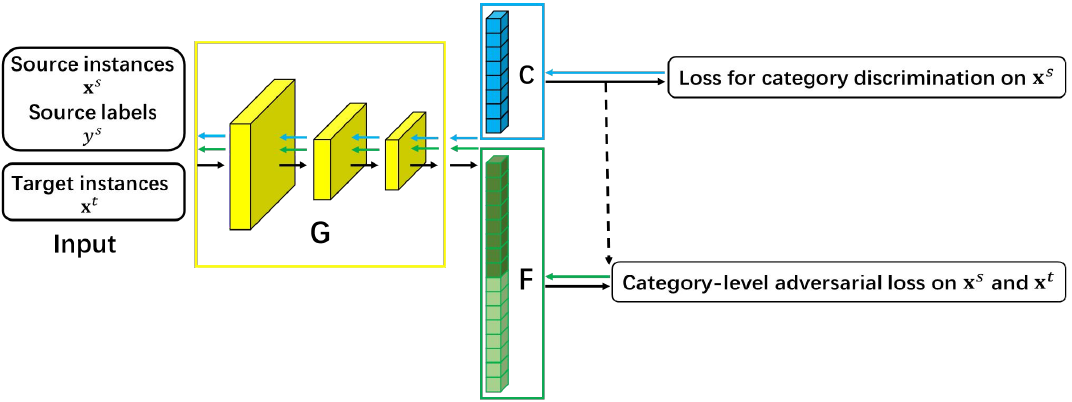}
    }

	\centering
	\subfigure[SymNet]{
		\label{fig:symnet}
		\includegraphics[width=0.85\linewidth]{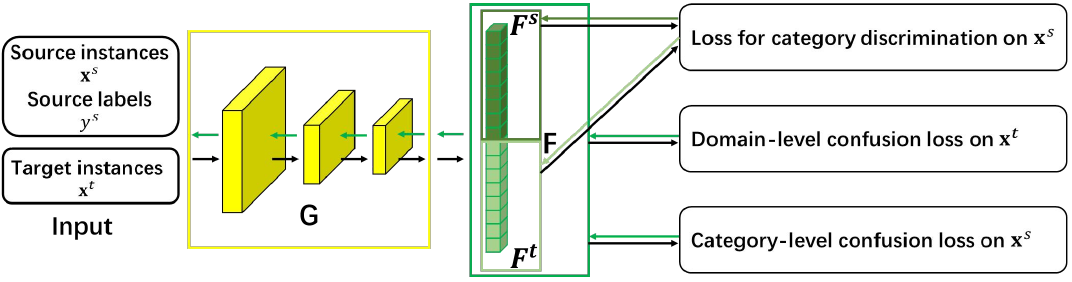}
	}

	\centering
	\subfigure[CatDA]{
		\label{fig:catda}
		\includegraphics[width=0.85\linewidth]{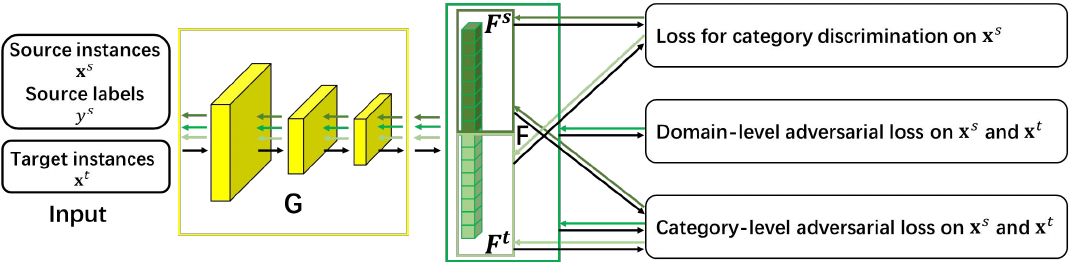}
	}

    \caption{Network architectures and loss designs of MADA, RCA, SymNet, and our CatDA. $G(\cdot)$ is the feature extractor. (a) MADA \cite{mada} includes a task classifier $C(\cdot)$ and multiple category-wise domain classifiers $\{F_k(\cdot)\}_{k=1}^K$. (b) RCA \cite{rca} contains a $C(\cdot)$ and a joint domain-category classifier $F(\cdot)$. (c) SymNet \cite{symnets} only comprises a $F(\cdot)$, but meanwhile considers it as two task classifiers of source $F^s(\cdot)$ and target $F^t(\cdot)$. (d) Our CatDA. Differently, CatDA based on adversarial training utilizes the target-discriminative information contained in category predictions from $F^s(\cdot)$ and $F^t(\cdot)$, i.e. the heterogenous, cross-domain weighting design.}
    \label{fig:net_arch}
    \end{figure}

	\paragraph{MADA \cite{mada}} Existing methods based on a single domain classifier \cite{dann,adda} disregard discriminative structures of data when aligning the two domains, resulting in the false alignment between different categories across domains. To reduce it, MADA uses multiple category-wise domain classifiers $\{F_k\}_{k=1}^K$, each of which takes as input the features weighted by the corresponding category prediction from the task classifier $C(\cdot)$ (see Fig. \ref{fig:mada}). Denote the cross-entropy loss as ${\cal{L}}_{\mathrm{ce}}(\cdot)$, the adversarial objective of MADA is 
	\begin{equation}
	\label{eqn:mada1}
	\min_{G, C} \frac{1}{n_s} \sum_{i=1}^{n_s} {\cal{L}}_{\mathrm{ce}}(C(G(\mathbf{x}_i^s)), y_i^s) - \lambda \frac{1}{n_s+n_t} \sum_{i=1}^{n_s+n_t} \sum_{k=1}^K {\cal{L}}_{\mathrm{ce}}(F_k( \hat{y}_{i,k} G(\mathbf{x}_i)), d_i), 
	\end{equation}
	\begin{equation}
	\label{eqn:mada2}
	\min_{\{F_k\}_{k=1}^K} \frac{1}{n_s+n_t} \sum_{i=1}^{n_s+n_t} \sum_{k=1}^K {\cal{L}}_{\mathrm{ce}}(F_k( \hat{y}_{i,k} G(\mathbf{x}_i)), d_i),
	\end{equation}
	where $\lambda$ is a hyper-parameter to trade-off the two loss terms in the unified optimization problem, $\hat{y}_{i,k}$ is the $k^{th}$ element of category prediction vector $\mathbf{\hat{y}}_i$ by $C(\cdot)$, and $d_i$ is the domain label for any instance $\mathbf{x}_i$, i.e. $0$ for the source domain and $1$ for the target one. This objective aligns each instance to the several most related categories, such that positive transfer can be promoted and negative transfer can be alleviated meanwhile.
	
	\paragraph{RCA \cite{rca}} To further reduce the false alignment, instead of a binary adversarial loss from a single domain classifier, 
	RCA imposes a $2K$-way adversarial loss from a joint domain-category classifier $F(\cdot)$ (see Fig. \ref{fig:rca}). The joint classifier considers the first $K$ as source categories and the last $K$ as target categories, and is learned by classifying any instance as its domain-category label, which naturally models a joint distribution over domain and category. Here, pseudo labels of unlabeled target instances are predicted by an additional task classifier $C(\cdot)$. Reversely, the feature extractor $G(\cdot)$ deceives $F(\cdot)$ by misclassifying any instance in terms of the domain label while keeping the category consistent. The adversarial objective of RCA is written as
	\begin{equation}
	\label{eqn:rca1}
	\min_{G, C} \frac{1}{n_s}\! \sum_{i=1}^{n_s}\! {\cal{L}}_{\mathrm{ce}}(C(G(\mathbf{x}_i^s)), y_i^s) + \lambda \left( \frac{1}{n_s}\! \sum_{i=1}^{n_s}\! {\cal{L}}_{\mathrm{ce}}(F(G(\mathbf{x}_i^s)), y_i^s\!+\! K) \!+\! \frac{1}{n_t}\! \sum_{j=1}^{n_t}\! {\cal{L}}_{\mathrm{ce}}(F(G(\mathbf{x}_j^t)), \hat{y}_j^t)\right),
	\end{equation}
	\begin{equation}
	\label{eqn:rca2}
	\min_{F} \frac{1}{n_s} \sum_{i=1}^{n_s} {\cal{L}}_{\mathrm{ce}}(F(G(\mathbf{x}_i^s)), y_i^s) + \frac{1}{n_t} \sum_{j=1}^{n_t} {\cal{L}}_{\mathrm{ce}}(F(G(\mathbf{x}_j^t), \hat{y}_j^t + K)),
	\end{equation}
    where $\hat{y}_j^t = \arg\max\limits_k C(G(\mathbf{x}_j^t))[k]$ is the predicted pseudo label by $C(\cdot)$. The joint classifier elegantly integrates the domain and category information, such that the domain alignment can be aware of category boundaries. On this basis, the above objective aims to learn invariant feature representations for instances from the same category of the two domains, which facilitates the alignment of class-conditional distributions across domains while forming disjoint supports for different categories in the feature space.
    
	\paragraph{SymNet \cite{symnets}} SymNet considers the joint classifier $F(\cdot)$ as two task classifiers of source $F^s(\cdot)$ and target $F^t(\cdot)$, and thus does not include an additional task classifier $C(\cdot)$ (see Fig. \ref{fig:symnet}). SymNet proposes the domain- and category-level confusion losses on target and source data respectively%, where the latter improves over the former by learning features to be invariant at the corresponding categories across domains
	, each of which computes the cross entropy between domain predictions and uniform distribution. The adversarial objective of SymNet is 
	\begin{eqnarray}
	\label{eqn:symnet1}
	\min_{F, F^s, F^t} \frac{1}{n_s} \sum_{i=1}^{n_s} {\cal{L}}_{\mathrm{ce}}(F^s(G(\mathbf{x}_i^s)), y_i^s) + \frac{1}{n_s} \sum_{i=1}^{n_s} {\cal{L}}_{\mathrm{ce}}(F^t(G(\mathbf{x}_i^s)), y_i^s) +\\ \notag \frac{1}{n_s+n_t} \sum_{i=1}^{n_s+n_t} {\cal{L}}_{\mathrm{ce}}(\left[\sum_{k=1}^K F(G(\mathbf{x}_i))[k], \sum_{k=1}^K F(G(\mathbf{x}_i))[k+K]\right], d_i), 
	\end{eqnarray}
	\begin{eqnarray}
	\label{eqn:symnet2}
	\min_{G} \frac{1}{2 n_s} \sum_{i=1}^{n_s} \left( {\cal{L}}_{\mathrm{ce}}(F(G(\mathbf{x}_i^s)), y_i^s) + {\cal{L}}_{\mathrm{ce}}(F(G(\mathbf{x}_i^s)), y_i^s + K)\right) + \\ \notag 
	%\lambda \frac{1}{2 n_t} \sum_{j=1}^{n_t} \left( {\cal{L}}_{\mathrm{ce}}(\left[\sum_{k=1}^K F(G(\mathbf{x}_j^t))[k], \sum_{k=1}^K F(G(\mathbf{x}_j^t))[k+K]\right], 0) + {\cal{L}}_{\mathrm{ce}}(\left[\sum_{k=1}^K F(G(\mathbf{x}_j^t))[k], \sum_{k=1}^K F(G(\mathbf{x}_j^t))[k+K]\right], 1) \right)
	\lambda \frac{1}{n_t} \sum_{j=1}^{n_t} - \left(0.5 \log \sum_{k=1}^K F(G(\mathbf{x}_j^t))[k] + 0.5 \log \sum_{k=1}^K F(G(\mathbf{x}_j^t))[k+K]\right),
	\end{eqnarray}
	where %$\mathbf{d}_i$ is $[1, 0]$ if $\mathbf{x}_i$ is a source instance and $[0, 1]$ otherwise, and 
	$F(\cdot)$ has a softmax layer in the top. In Eq. (\ref{eqn:symnet2}), the first term on $\mathbf{x}^s$ and the second term on $\mathbf{x}^t$ are for the domain- and category-level domain confusions respectively, aiming to align the joint distributions of feature and category across domains. However, such a domain alignment is sub-optimal since each level of domain confusion is performed on one domain only; especially, SymNet ignores the target-discriminative information contained in category predictions from $F^s(\cdot)$ and $F^t(\cdot)$.%, just like RCA.
	
	\subsection{Categorical Domain Adaptation}
	\label{subsec:32}
	
	Motivated to address the above issues, based on the joint classifier $F(\cdot)$ concatenating the source and target task classifiers of $F^s(\cdot)$ and $F^t(\cdot)$, we propose to apply novel losses of adversarial training \emph{at multiple levels} to probabilities of softmax outputs of $F(\cdot)$, $F^s(\cdot)$, and $F^t(\cdot)$, to promote \emph{categorical domain adaptation (CatDA)}. An important technique wherein is a design of cross-entropy losses concerning probability interactions between source and target category predictions.
	
	Given an input instance $\mathbf{x}$, denote the probability vectors of the softmax outputs of $F(G(\mathbf{x}))$, $F^s(G(\mathbf{x}))$, and $F^t(G(\mathbf{x}))$ respectively as
	\begin{align}\label{eqn:compute_prob}
	\mathbf{p}(\mathbf{x})  = F(G(\mathbf{x})) \in [0, 1]^{2K} ; 
	\mathbf{p}^s(\mathbf{x})  = F^s(G(\mathbf{x})), 
	\mathbf{p}^t(\mathbf{x})  = F^t(G(\mathbf{x})) \in [0, 1]^K. 
	\end{align}
	For ease of notations, we also write $p_k(\mathbf{x})$ (\emph{resp.} $p_k^s(\mathbf{x})$ or $p_k^t(\mathbf{x})$) for the $k^{th}$ element of the probability vector $\mathbf{p}(\mathbf{x})$ (\emph{resp.} $\mathbf{p}^s(\mathbf{x})$ or $\mathbf{p}^t(\mathbf{x})$). We use $\mathbf{p}(\mathbf{x})$, $\mathbf{p}^s(\mathbf{x})$, and $\mathbf{p}^t(\mathbf{x})$ to define our proposed adversarial losses, and also the loss for task classification. Adversarial training aims to learn a domain-invariant feature extractor $G(\cdot)$, and also $F^s(\cdot)$ and $F^t(\cdot)$, which share network parameters with $F(\cdot)$.
	
	\paragraph{Loss for Category Discrimination} For our defined source and target task classifiers $F^s(\cdot)$ and $F^t(\cdot)$, it is natural to expect their category predictions are corresponded. In other words, for an instance $\mathbf{x}$ of the $k^{th}$ category,  both $p_k^s(\mathbf{x})$ and $p_k^t(\mathbf{x})$ are expected to be the elements of the highest scores respectively in $\mathbf{p}^s(\mathbf{x})$ and $\mathbf{p}^t(\mathbf{x})$. Since only source instances are labeled, to achieve the above effect, we use $\{ (\mathbf{x}_i^s, y_i^s) \}_{i=1}^{n_s}$ to simultaneously train $F^s(\cdot)$ and $F^t(\cdot)$ with 
	\begin{equation}\label{eqn:src_cls_loss}
	{\cal{L}}_{\mathrm{cls}}(G, F) = {\cal{L}}_{\mathrm{cls}}(G, F^s, F^t) = - \frac{1}{n_s} \sum_{i=1}^{n_s} \log p_{y_i^s}^s(\mathbf{x}_i^s) - \frac{1}{n_s} \sum_{i=1}^{n_s} \log p_{y_i^s}^t(\mathbf{x}_i^s).
	\end{equation}
	The classification loss of Eq. (\ref{eqn:src_cls_loss}) will be used together with the domain- and category-level adversarial losses to constitute our objective of CatDA, as explained shortly.
	
	\paragraph{Domain-Level Adversarial Loss} As illustrated in Fig. \ref{fig:catda}, our used network has no an explicit domain classifier. To define a domain-level adversarial loss, we take the first $K$ neurons of the last layer of $F(\cdot)$ \emph{collectively} as the source domain, and its last $K$ neurons \emph{collectively} as the target domain. We accordingly sum up the element probabilities $\sum_{k=1}^{K} p_k(\mathbf{x})$ and $\sum_{k=1}^{K} p_{k+K}(\mathbf{x})$, and use standard binary cross entropy to define our domain-level adversarial loss as 
	\begin{eqnarray}\label{eqn:dom_adv_loss}
	\begin{aligned}
	& \min_{F} {\cal{F}}_{\mathrm{adv}}^D(G, F) = - \frac{1}{n_s} \sum_{i=1}^{n_s} {\log \sum_{k=1}^{K} p_k(\mathbf{x}_i^s)} - \frac{1}{n_t} \sum_{j=1}^{n_t} {\log \sum_{k=1}^{K} p_{k+K}(\mathbf{x}_j^t)}, \\
	& \min_{G} {\cal{G}}_{\mathrm{adv}}^D(G, F) = - \frac{1}{n_s} \sum_{i=1}^{n_s} {\log \sum_{k=1}^{K} p_{k+K}(\mathbf{x}_i^s)} - \frac{1}{n_t} \sum_{j=1}^{n_t} {\log \sum_{k=1}^{K} p_k(\mathbf{x}_j^t)},
	\end{aligned}
	\end{eqnarray}
	where we take the inverted (domain) label version of minimax loss \cite{gans,adda} to address the issue of vanishing gradients, which splits the adversarial loss into two independent ones to update network parameters involved in $F(\cdot)$ and $G(\cdot)$ respectively.
	
	\paragraph{Category-Level Adversarial Loss} We have expected that the first and last $K$ neurons of the last layer of $F(\cdot)$ (i.e. neurons of the respective last layers of $F^s(\cdot)$ and $F^t(\cdot)$) are corresponded in terms of category predictions, which means that for any instance $\mathbf{x}$ from the two domains, $p_y(\mathbf{x})$ and $p_{y+K}(\mathbf{x})$ are of the highest probabilities (correspondingly, $p_y^s(\mathbf{x})$ and $p_y^t(\mathbf{x})$ are respectively of the highest probabilities in $\mathbf{p}^s(\mathbf{x})$ and $\mathbf{p}^t(\mathbf{x})$). We try to enforce this expectation by simultaneously training $F^s(\cdot)$ and $F^t(\cdot)$ using the classification loss of Eq. (\ref{eqn:src_cls_loss}) on $\{ (\mathbf{x}_i^s, y_i^s) \}_{i=1}^{n_s}$. Our category-level adversarial loss defined over $F(\cdot)$ is based on this expectation as well.
	
	A challenge still remains to implement category-level adversarial training on both $\{ (\mathbf{x}_i^s, y_i^s) \}_{i=1}^{n_s}$ and $\{ \mathbf{x}_j^t \}_{j=1}^{n_t}$, since for any target instance $\mathbf{x}^t$, we have no a reliable way to specify its labeling. To address it, we propose a \emph{heterogenous, cross-domain weighting scheme} to aggregate category probability predictions of $\mathbf{x}^t$ over all the $K$ categories as a proxy of its labeling, where weighting factors themselves are predicted pseudo labels (probabilities) from the task classifiers $F^s(\cdot)$ or $F^t(\cdot)$. Based on the scheme, we again use cross entropy to define our category-level adversarial loss as 
	\begin{eqnarray}\label{eqn:cat_adv_loss}
	\begin{aligned}
	& \min_{F} {\cal{F}}_{\mathrm{adv}}^C(G,\! F) \!=\! {\cal{F}}_{\mathrm{adv}}^C(G,\! F,\! F^s) \!=\! - \frac{1}{n_s}\! \sum_{i=1}^{n_s} \log p_{y_i^s}(\mathbf{x}^s_i)\! -\! \frac{1}{n_t}\! \sum_{j=1}^{n_t} \sum_{k=1}^K p_k^s(\mathbf{x}^t_j) \log p_{k+K}(\mathbf{x}^t_j), \\
	& \min_{G} {\cal{G}}_{\mathrm{adv}}^C(G,\! F) \!=\! {\cal{G}}_{\mathrm{adv}}^C(G,\! F,\! F^t) \!=\! - \frac{1}{n_s}\! \sum_{i=1}^{n_s} \log p_{y_i^s+K}(\mathbf{x}^s_i)\! -\! \frac{1}{n_t}\! \sum_{j=1}^{n_t} \sum_{k=1}^K p_k^t(\mathbf{x}^t_j) \log p_{k}(\mathbf{x}^t_j), 
	\end{aligned}
	\end{eqnarray}
	where the product terms involving pseudo labels and log probabilities further enhance the consistency of category predictions for target instances between $F^s(\cdot)$ and $F^t(\cdot)$, since the two task classifiers are defined by the same FC layer of the concatenated classifier $F(\cdot)$. The use of pseudo label predictions from the task classifier of one domain to guide category probability predictions of $F(\cdot)$ on another domain (i.e. the heterogenous, cross-domain weighting scheme) also improves the reliability of CatDA in the early stage of training (cf. Fig. \ref{fig:catda_cdw_vs_sdw} for experimental evidence). 
	
	\noindent\textbf{Remark.} Since categorically corresponded discriminative training of $F^s(\cdot)$ and $F^t(\cdot)$ are continuously enforced using the classification loss of Eq. (\ref{eqn:src_cls_loss}) over the labeled source data, pseudo label predictions of the target data from $F^s(\cdot)$ or $F^t(\cdot)$ will remain wrong in the subsequent stage of training if they are wrong in the early stage of training. These wrong pseudo label predictions will have no chance to be corrected if pseudo label predictions from the task classifier of one domain are used to guide category probability predictions of $F(\cdot)$ on the same domain. Conversely, these wrong pseudo label predictions will have a chance to be corrected if the proposed cross-domain weighting scheme is used, since for any target instance, pseudo label predictions from the task classifier of one domain could be right when those from the task classifier of another domain are wrong, i.e. $F^s(\mathbf{x}^t)$ and $F^t(\mathbf{x}^t)$ could be complementary. 
	Also because of Eq. (\ref{eqn:src_cls_loss}), pseudo label predictions of the target data from $F^s(\cdot)$ or $F^t(\cdot)$ that are right in the early stage of training, are hard to be wrong in the subsequent stage of training. The above analysis explains the effectiveness of our proposed scheme.
	
	\paragraph{Overall Objective of CatDA} Combining the loss of Eq. (\ref{eqn:src_cls_loss}) for task classification, and the domain- and category-level adversarial losses of Eq. (\ref{eqn:dom_adv_loss}) and Eq. (\ref{eqn:cat_adv_loss}) gives the overall training objective of CatDA as 
	\begin{eqnarray}\label{eqn:catda}
	\begin{aligned}
	& \min_{F} {\cal{L}}_{\mathrm{cls}}(G, F) + {\cal{F}}_{\mathrm{adv}}^D(G, F) + \lambda {\cal{F}}_{\mathrm{adv}}^C(G, F),  \\
	& \min_{G} \frac{1}{2}{\cal{L}}_{\mathrm{cls}}(G, F) + \lambda {\cal{G}}_{\mathrm{adv}}^D(G, F) + \lambda {\cal{G}}_{\mathrm{adv}}^C(G, F),
	\end{aligned}
	\end{eqnarray}
	where we use Eq. (\ref{eqn:src_cls_loss}) to update $G(\cdot)$ as well. This is to ensure that during adversarial training, the respective category discrimination of $F^s(\cdot)$ and $F^t(\cdot)$ could be maintained, which is neglected in SymNet. When minimizing over $G(\cdot)$, we halve the loss ${\cal{L}}_{\mathrm{cls}}$ using a factor of $1/2$ to normalize two flows of back-propagated gradients onto $G(\cdot)$, which are respectively from $F^s(\cdot)$ and $F^t(\cdot)$ (i.e. the two terms in Eq. (\ref{eqn:src_cls_loss})). Fig. \ref{fig:catda} gives an illustration. We also use a penalty $\lambda$ that is progressively increased from $0$ to $1$ (cf. Section \ref{subsec:52} for its rule of equation), to suppress signals from a few terms of Eq. (\ref{eqn:catda}), which could be less reliable in the early stage of training: $\lambda$ before ${\cal{F}}_{\mathrm{adv}}^C$ and ${\cal{G}}_{\mathrm{adv}}^C$ is for pseudo labels of the target data from $F^s(\cdot)$ and $F^t(\cdot)$, and $\lambda$ before ${\cal{G}}_{\mathrm{adv}}^D$ is to reduce the false alignment between different categories across the two domains.
	
	\subsection{Enhancement with Vicinal Domain Adaptation}
	\label{subsec:33}
	
	\begin{figure}
		\centering		
		\subfigure[]{
			\label{fig:vicinal_domains}
			\includegraphics[width=0.6\linewidth, height=0.25\linewidth]{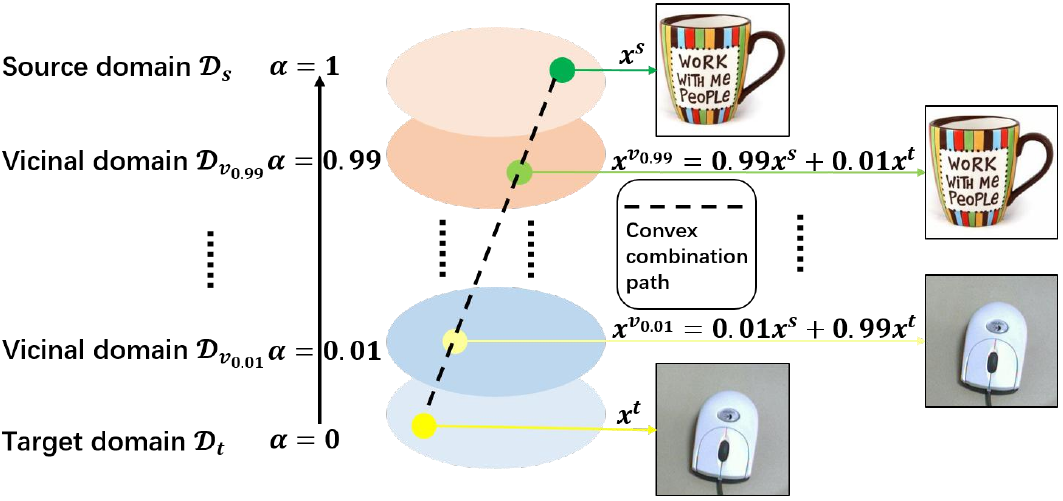}
		}	
		%\hspace{0.1in}		
		\subfigure[]{
			\label{fig:beta_0p2}
			\includegraphics[width=0.35\linewidth, height=0.25\linewidth]{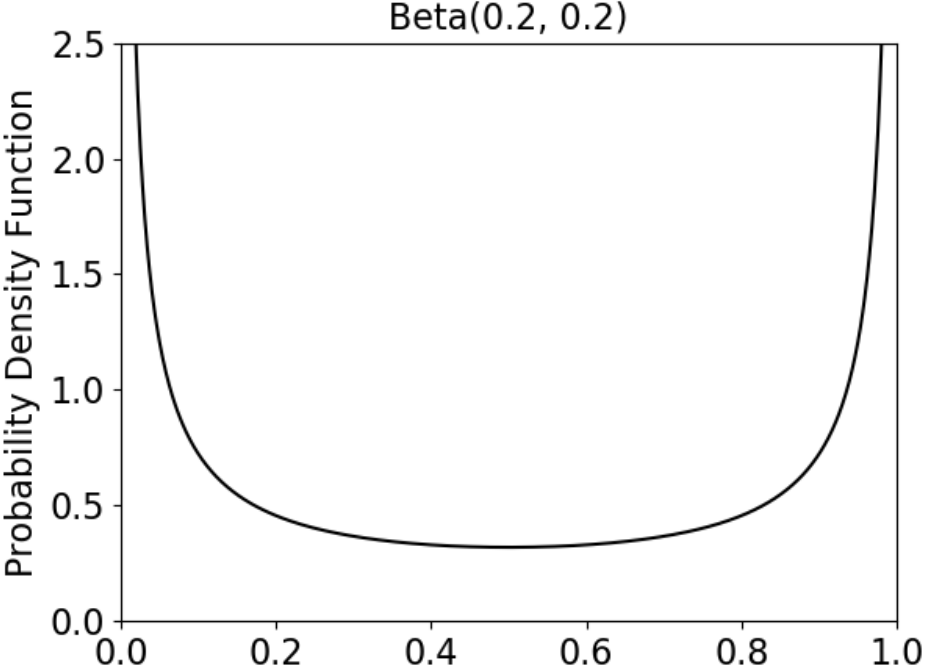}
		}		
		\caption{(a) An example to illustrate how instances of vicinal domains are synthesized. Domains are denoted by ovals. Instances are denoted by circles. Intuitively, if the corresponding categories in the vicinal domains of ${\cal{D}}_{v_{\alpha}}$ and ${\cal{D}}_{v_{1-\alpha}}$ are well aligned, they will be well aligned in the source and target domains of ${\cal{D}}_s$ and ${\cal{D}}_t$. Note that the source and target domains correspond to \textbf{A} and \textbf{W} of the Office-31 \cite{office31} dataset respectively. Please refer to the appendix for more examples on other three datasets. (b) $\alpha \sim {\rm Beta}(\beta, \beta)$ with $\beta=0.2$. }
		\label{fig:vicinal_domain_adaptation}
	\end{figure}
	
	Most of the existing methods pursue domain adaptation of the given ${\cal{D}}_s$ and ${\cal{D}}_t$ themselves. There exists a vertical direction to pursue that generates (statistics or features) of augmented domains from ${\cal{D}}_s$ and ${\cal{D}}_t$, and improves domain adaptation by leveraging these augmented domains \cite{gfk,dlow}. We are also motivated to contribute to this direction that is not well-studied yet. Differently, we are inspired by the work \cite{mixup} and plainly generate augmented domains in the vicinities of ${\cal{D}}_s$ and ${\cal{D}}_t$ (i.e. the vicinal domains), whose instances are the convex combination of pairs of instances respectively from ${\cal{D}}_s$ and ${\cal{D}}_t$, as illustrated in Fig. \ref{fig:vicinal_domains}. Intuitively, if ${\cal{D}}_s$ and ${\cal{D}}_t$ are perfectly aligned, the generated vicinal domains are aligned as well. It is thus a sensible way to align vicinal domains to improve the alignment of ${\cal{D}}_s$ and ${\cal{D}}_t$. We propose novel adversarial losses for vicinal domain adaptation (VicDA) based on CatDA. We term our full version method as \emph{Vicinal and Categorical Domain Adaptation (ViCatDA)}.

	Technically, given $\alpha \in [0, 1]$, we denote a vicinal domain as ${\cal{D}}_{v_{\alpha}}=\{\mathbf{x}_l^{v_{\alpha}}\}_{l=1}^{n_{v_{\alpha}}}$, where $n_{v_{\alpha}}$ is the number of instances on ${\cal{D}}_{v_{\alpha}}$. 
	The instance $\mathbf{x}^{v_{\alpha}}$ is generated by a convex combination of randomly sampled $\mathbf{x}^s \in {\cal{D}}_s$ and $\mathbf{x}^t \in {\cal{D}}_t$ as 
	\begin{eqnarray}\label{eqn:ins_vic_dom}
	\begin{aligned}
	\mathbf{x}^{v_{\alpha}} = \alpha \mathbf{x}^s + (1 - \alpha) \mathbf{x}^t .
	\end{aligned}
	\end{eqnarray}
	We follow mixup \cite{mixup} to sample $\alpha$ from a beta distribution ${\rm Beta}(\beta, \beta)$ with $\beta=0.2$ (cf. Fig. \ref{fig:beta_0p2}), which means that a theoretically infinite number of vicinal domains $\{ {\cal{D}}_{v_{\alpha}} \}$ can be generated. Assuming a total of $n_v$ instances are generated for $\{ {\cal{D}}_{v_{\alpha}} \}$, we propose the VicDA version of domain-level adversarial loss that extends the loss in Eq. (\ref{eqn:dom_adv_loss}) as 
	\begin{eqnarray}\label{eqn:vic_dom_adv_loss}
	\begin{aligned}
	& \min_{F} {\cal{F}}_{\mathrm{adv}}^{VD}(G, F) = -\frac{1}{n_v} \sum_{l=1}^{n_v} \left( \alpha \log \sum_{k=1}^{K} p_k(\mathbf{x}_l^{v_{\alpha}}) + (1 - \alpha) \log \sum_{k=1}^K p_{k+K}(\mathbf{x}_l^{v_{\alpha}})\right), \\
	& \min_{G} {\cal{G}}_{\mathrm{adv}}^{VD}(G, F) = -\frac{1}{n_v} \sum_{l=1}^{n_v} \left( \alpha \log \sum_{k=1}^{K} p_{k+K}(\mathbf{x}_l^{v_{\alpha}}) + (1 - \alpha) \log \sum_{k=1}^K p_k(\mathbf{x}_l^{v_{\alpha}})\right), 
	\end{aligned}
	\end{eqnarray}
	where for any instance $\mathbf{x}^{v_{\alpha}} \in {\cal{D}}_{v_{\alpha}}$, the log of collective probabilities over either the first or the last $K$ neurons of $F(\cdot)$ is weighted by $\alpha$ to enforce the labeling of vicinal domain ${\cal{D}}_{v_{\alpha}}$. We similarly propose our VicDA version of category-level adversarial loss by extending the loss in Eq. (\ref{eqn:cat_adv_loss}) as 
	\begin{eqnarray}\label{eqn:vic_cat_adv_loss}
	\begin{aligned}
	& \min_{F} {\cal{F}}_{\mathrm{adv}}^{VC}(G, F) = -\frac{1}{n_v}  \sum_{l=1}^{n_v} \Bigg(\alpha \log p_{y^s}(\mathbf{x}_l^{v_{\alpha}}) + (1 - \alpha) \sum_{k=1}^K p_k^s(\mathbf{x}_l^t) \log p_{k+K}(\mathbf{x}_l^{v_{\alpha}})\Bigg), \\
	& \min_{G} {\cal{G}}_{\mathrm{adv}}^{VC}(G, F) = -\frac{1}{n_v}  \sum_{l=1}^{n_v} \Bigg( \alpha \log p_{y^s+K}(\mathbf{x}_l^{v_{\alpha}}) + (1 - \alpha) \sum_{k=1}^K p_k^t(\mathbf{x}_l^t) \log p_{k}(\mathbf{x}_l^{v_{\alpha}}) \Bigg),
	\end{aligned}
	\end{eqnarray}
	where $\mathbf{x}_l^t$ with $l \in \{1, \dots, n_v\}$ denotes the target instance that generates $\mathbf{x}_l^{v_{\alpha}}$ by Eq. (\ref{eqn:ins_vic_dom}). 
	Replacing the terms of ${\cal{F}}_{\mathrm{adv}}^D$, ${\cal{F}}_{\mathrm{adv}}^C$, ${\cal{G}}_{\mathrm{adv}}^D$, and ${\cal{G}}_{\mathrm{adv}}^C$ in the overall objective of Eq. (\ref{eqn:catda}) of CatDA with the respective VicDA versions of ${\cal{F}}_{\mathrm{adv}}^{VD}$, ${\cal{F}}_{\mathrm{adv}}^{VC}$, ${\cal{G}}_{\mathrm{adv}}^{VD}$, and ${\cal{G}}_{\mathrm{adv}}^{VC}$ gives our overall objective of ViCatDA 
	\begin{eqnarray}\label{eqn:vicatda}
	\begin{aligned}
	& \min_{F} {\cal{L}}_{\mathrm{cls}}(G, F) + {\cal{F}}_{\mathrm{adv}}^{VD}(G, F) + \lambda {\cal{F}}_{\mathrm{adv}}^{VC}(G, F),  \\
	& \min_{G} \frac{1}{2}{\cal{L}}_{\mathrm{cls}}(G, F) + \lambda {\cal{G}}_{\mathrm{adv}}^{VD}(G, F) + \lambda {\cal{G}}_{\mathrm{adv}}^{VC}(G, F). 
	\end{aligned}
	\end{eqnarray}
	ViCatDA can stabilize adversarial training by regularizing gradients of $F(\cdot)$, $F^s(\cdot)$, and $F^t(\cdot)$, leading to a stable source of gradient information to $G(\cdot)$ \cite{mixup}. Experiments show the efficacy of enhancing CatDA with VicDA to have ViCatDA. For more clarity, we summarize the main steps of the training process of ViCatDA in the appendix% Algorithm \ref{alg:vicatda}
	. 
	
	%\subsection{Training Algorithm}	

	\subsection{Target Discriminative Structure Recovery}
	\label{subsec:34}
	
	Motivated by a fact that the adversarial feature alignment could damage the intrinsic discriminative structures of unlabeled target data, as discussed in recent works \cite{bsp,tat,da_theory3}. In this work, we also propose \emph{Target Discriminative Structure Recovery (TDSR)} to recover the damaged target discriminative structures, via further fine-tuning the trained ViCatDA model on unlabeled target samples with cluster labels assigned by the established data clustering technique \cite{spherical_k_means,cluster_review}. Data clustering is to group unlabeled instances into semantically meaningful clusters in a data-driven way, i.e. discovering the intrinsic discriminative structures of unlabeled data. Among various clustering algorithms, we choose the simple but flexible spherical k-means \cite{spherical_k_means}, which enjoys both good solution quality and high computational efficiency. Specifically, it uses the cosine dissimilarity to perform prototype-based data partitioning. We term this algorithm as semantically anchored spherical k-means in that we use a trained ViCatDA as the initial clustering model, which has already contained rich semantic information. 
	
	At each training epoch, we first estimate the class label $\hat{y}_j^t$ of each target sample $\mathbf{x}_j^t$ in the maximum likelihood principle, i.e. taking the class with the highest probability predicted by the target task classifier $F^t(\cdot)$, 
	and obtain the cluster center ${\mathbf{\mu}}_k^t$
	by summing the $l_2$ normalized feature vectors of target samples
	with same category prediction as
	\begin{equation}	
	{\mathbf{\mu}}_k^t = \sum_{j=1}^{n_k^t} \frac{G(\mathbf{x}_j^t)}{\|G(\mathbf{x}_j^t)\|_2}, \;\; {\rm s.t.} \;\; \hat{y}_j^t = k
	\label{eq:target_center_init}
	\end{equation}
	where $n_k^t$ denotes the number of target samples predicted as the $k^{th}$ class.
	Starting with semantically anchored center initialization (Eq. (\ref{eq:target_center_init})), 
	we refine clusters by alternating the following two steps.
	{\bf 1)} Updating the cluster assignment of each $\mathbf{x}_j^t$ based on the minimum cosine dissimilarity principle, i.e. $\hat{y}_j^t=\arg\min\limits_k \frac{1}{2} (1 - \frac{\langle G(\mathbf{x}_j^t), \; {\mathbf{\mu}}_k^t \rangle}{\| G(\mathbf{x}_j^t) \|_2 \; \| {\mathbf{\mu}}_k^t \|_2})$.
	{\bf 2)} Updating the cluster center based on the new cluster assignments with Eq. (\ref{eq:target_center_init}).
	This process repeats until no change in cluster assignments. 
	Then, based on the final cluster assignments of target samples, we optimize the objective of TDSR as 
	\begin{eqnarray}\label{EqnTDSR}
	\begin{aligned}
	\min_{G, F^t} {\cal{L}}_{tdsr} (G, F^t) = -\frac{1}{n_t} \sum_{j=1}^{n_t} \log p_{{\hat{y}}_j^t}^t(\mathbf{x}_j^t).
	\end{aligned}
	\end{eqnarray}
	
	\section{Method Analysis}
	\label{sec:analysis}
	
	\subsection{Cross-Domain Weighting Scheme}
	\label{subsec:41}
	
	In this section, we analyze our proposed cross-domain weighting scheme from the perspective of information theory. 

	Let $\mathbf{p}_{1:K}(\mathbf{x})$ and $\mathbf{p}_{K+1:2K}(\mathbf{x})$ be the subvectors containing the first $K$ and last $K$ elements of $\mathbf{p}(\mathbf{x})$ respectively. Let $D_{KL}(\mathbf{p}||\mathbf{q})$ be the Kullback-Leibler (KL) divergence between two probability vectors $\mathbf{p}$ and $\mathbf{q}$, which are defined on the same probability space. Let $H(\mathbf{p})$ be the information entropy of a probability vector $\mathbf{p}$. When minimized over the joint classifier $F(\cdot)$, our proposed category-level target adversarial loss is written as 
	\begin{eqnarray}\label{eqn:cat_adv_loss_F_tar}
	\begin{aligned}
	{\cal{F}}_{\textrm{t-adv}}^C &= - \frac{1}{n_t} \sum_{j=1}^{n_t} \sum_{k=1}^K p_k^s(\mathbf{x}^t_j) \log p_{k+K}(\mathbf{x}^t_j) \\
	&= - \frac{1}{n_t} \sum_{j=1}^{n_t} \sum_{k=1}^K p_k^s(\mathbf{x}^t_j) \log \frac{p_{k+K}(\mathbf{x}^t_j)}{p_k^s(\mathbf{x}^t_j) } - \frac{1}{n_t} \sum_{j=1}^{n_t} \sum_{k=1}^K p_k^s(\mathbf{x}^t_j) \log p_k^s(\mathbf{x}^t_j) \\
	&= \frac{1}{n_t} \sum_{j=1}^{n_t} D_{KL}(\mathbf{p}^s(\mathbf{x}_j^t) || \mathbf{p}_{K+1:2K}(\mathbf{x}_j^t)) + \frac{1}{n_t} \sum_{j=1}^{n_t} H(\mathbf{p}^s(\mathbf{x}_j^t)). 
	\end{aligned}
	\end{eqnarray}
	Similarly, when minimized over the feature extractor $G(\cdot)$, our proposed category-level target adversarial loss is written as 
	\begin{eqnarray}\label{eqn:cat_adv_loss_G_tar}
	\begin{aligned}
	{\cal{G}}_{\textrm{t-adv}}^C & = \frac{1}{n_t} \sum_{j=1}^{n_t} D_{KL}(\mathbf{p}^t(\mathbf{x}_j^t) || \mathbf{p}_{1:K}(\mathbf{x}_j^t)) + \frac{1}{n_t} \sum_{j=1}^{n_t} H(\mathbf{p}^t(\mathbf{x}_j^t)). 
	\end{aligned}
	\end{eqnarray}
	Minimizing ${\cal{F}}_{\textrm{t-adv}}^C$ over $F(\cdot)$ is equivalent to reduce the KL-divergence between $\mathbf{p}^s(\mathbf{x}_j^t)$ and $\mathbf{p}_{K+1:2K}(\mathbf{x}_j^t)$ while decreasing the entropy of $\mathbf{p}^s(\mathbf{x}_j^t)$; minimizing ${\cal{G}}_{\textrm{t-adv}}^C$ over $G(\cdot)$ is equivalent to reduce the KL-divergence between $\mathbf{p}^t(\mathbf{x}_j^t)$ and $\mathbf{p}_{1:K}(\mathbf{x}_j^t)$ while decreasing the entropy of $\mathbf{p}^t(\mathbf{x}_j^t)$. Such an adversarial optimization suggests that $F(\cdot)$ is fully confused, i.e. reaching an equilibrium in the two-player game \cite{GenEqu}, only when the probability vectors $\mathbf{p}_{1:K}(\mathbf{x}^t)$ and $\mathbf{p}_{K+1:2K}(\mathbf{x}^t)$ are the same as a unimodal distribution with the maximum value of $0.5$ for any target instance $\mathbf{x}^t$. %Hence, the indexes of the maximum values of $\mathbf{p}_{1:K}(\mathbf{x}^t)$ and $\mathbf{p}_{K+1:2K}(\mathbf{x}^t)$ are the same. 
	The optimized result manifests the complete consistency of category predictions between $F^s(\cdot)$ and $F^t(\cdot)$ in terms of both the predicted category label and prediction confidence. If pseudo label predictions from the task classifier of one domain are used to guide category probability predictions of $F(\cdot)$ on this domain, category predictions between $F^s(\cdot)$ and $F^t(\cdot)$ may be inconsistent in terms of the predicted category label.
	
	These analyses concretize the theoretical result of labeling consistency across domains \cite{da_theory2,da_theory1} and further explain the effectiveness of our cross-domain weighting scheme.
	
	\subsection{Multi-Level Adversarial Training}
	\label{subsec:42}
	
	In this section, we provide more explanations for the loss in Eq. (\ref{eqn:cat_adv_loss}) and the relation between the losses in Eq. (\ref{eqn:dom_adv_loss}) and Eq. (\ref{eqn:cat_adv_loss}).
	
	The motivation of our category-level adversarial loss of Eq. (\ref{eqn:cat_adv_loss}) is to achieve category-level alignment of features and classifiers across domains. Minimizing Eq. (\ref{eqn:cat_adv_loss}) over the joint classifier $F(\cdot)$ approaches optimal solutions of $\mathbf{p}(\mathbf{x}^s)=[{\mathbf{y}^s}^T, \mathbf{0}^T]^T$ for any $(\mathbf{x}^s, \mathbf{y}^s)$ where $\mathbf{y}^s$ is $K$-dimensional one-hot label of $\mathbf{x}^s$ and $\mathbf{0}$ is a $K$-dimensional all-zero vector, and $\mathbf{p}(\mathbf{x}^t)=[\mathbf{0}^T, \mathbf{p}^s(\mathbf{x}^t )^T]^T$ for any $\mathbf{x}^t$, by adapting decision boundaries so that the task classifier of one domain can distinguish categories of instances of this domain; minimizing Eq. (\ref{eqn:cat_adv_loss}) over the feature extractor $G(\cdot)$ approaches those of $\mathbf{p}(\mathbf{x}^s)=[\mathbf{0}^T, {\mathbf{y}^s}^T]^T$ and $\mathbf{p}(\mathbf{x}^t )=[\mathbf{p}^t(\mathbf{x}^t )^T, \mathbf{0}^T]^T$ by learning features so that the task classifier of one domain can discriminate categories of instances of another domain. 
	
	Minimizing Eq. (\ref{eqn:dom_adv_loss}) over $F(\cdot)$ approaches optimal solutions of $\sum_{k=1}^K  p_k(\mathbf{x}^s)=1$ and $\sum_{k=1}^K  p_{k+K}(\mathbf{x}^s)=0$, and $\sum_{k=1}^K  p_k(\mathbf{x}^t)=0$ and $\sum_{k=1}^K  p_{k+K}(\mathbf{x}^t)=1$; minimizing Eq. (\ref{eqn:dom_adv_loss}) over $G(\cdot)$ approaches those of $\sum_{k=1}^K  p_k(\mathbf{x}^s)=0$ and $\sum_{k=1}^K  p_{k+K}(\mathbf{x}^s)=1$, and $\sum_{k=1}^K  p_k(\mathbf{x}^t)=1$ and $\sum_{k=1}^K  p_{k+K}(\mathbf{x}^t)=0$. Intuitively, Eq. (\ref{eqn:dom_adv_loss}) tries to classify any instance to either source or target domain by $F(\cdot)$, and Eq. (\ref{eqn:cat_adv_loss}) tries to identify its (pseudo) category label by the task classifier of this domain. Eq. (\ref{eqn:cat_adv_loss}) improves over Eq. (\ref{eqn:dom_adv_loss}) by driving domain-adversarial training from the domain to category level, i.e. competing between the corresponding categories of the source and target domains.
	
	\subsection{Vicinal Domain Adaptation}
	\label{subsec:43}
	
	In this section, we further clarify our proposed vicinal domain adaptation (VicDA).
	
	VicDA produces instances of vicinal domains via a convex combination of pairs of raw instances (e.g. images) respectively from the source and target domains, as shown in Fig. \ref{fig:vicinal_domains}. It may be less intuitive to train a model using such virtual instances, especially when raw instances are of different categories or under varying imaging conditions, however, one can understand this strategy as data augmentation that extends the benign behavior of trained model linearly between instances \cite{mixup}. This strategy is used in mixup \cite{mixup} for supervised learning on a single domain; we extend it for multi-level adversarial domain adaptation. Note that our VicDA formulation in the losses of Eq. (\ref{eqn:vic_dom_adv_loss}) and Eq. (\ref{eqn:vic_cat_adv_loss}) is not based on target pseudo labels, where $\alpha$ is simply the variable indicating the closeness of a virtual instance to the source or target domains. The smaller $\alpha$, the closer a virtual instance is to the target domain. Thus, we should impose less force on its feature alignment to the target domain, such that the originally well-aligned categories between the source and target domains will not be incorrectly mapped \cite{clan}. This is the essential cause of weighting the terms of aligning a virtual instance to the target domain in Eq. (\ref{eqn:vic_dom_adv_loss}) and Eq. (\ref{eqn:vic_cat_adv_loss}) by $\alpha$. While analysis on the theoretical stability of VicDA may be pursued, our empirical results have already confirmed its efficacy.

\section{Experiments}
\label{sec:experiments}

\subsection{Datasets}
\label{subsec:51}

\textbf{Office-31} \cite{office31} is a popular benchmark dataset for visual domain adaptation, which contains $4,110$ images of $31$ object categories shared by three domains: Amazon (\textbf{A}), Webcam (\textbf{W}), and DSLR (\textbf{D}). We evaluate on all the $6$ adaptation tasks.

\textbf{Office-Home} \cite{officehome} is a much more challenging benchmark dataset, which includes about $15,500$ images of $65$ object categories shared by four extremely distinct domains: Artistic images (\textbf{Ar}), Clip Art (\textbf{Cl}), Product images (\textbf{Pr}), and Real-World images (\textbf{Rw}). We evaluate on all the $12$ adaptation tasks.

\textbf{VisDA-2017} \cite{visda2017} is a difficult simulation-to-real benchmark. There are over
$280$K images of $12$ categories shared by: Training (\textbf{Synthetic}), Validation (\textbf{Real}), and Testing. Images of the domain Training are collected by rendering $3$D models and the other two domains comprise real-world images. We evaluate on the \textbf{Synthetic}$\rightarrow$\textbf{Real} task.

%\begin{table}[!t]
%	\centering
%	\caption{Dataset statistics.}
%	\label{table:dataset_statistics}
%	%\vskip -0.5cm
%	\resizebox{1.0\linewidth}{!}{
%		\begin{tabular}{c|ccc|cccc|cc|ccc}
%			\toprule
%			Datasets & \multicolumn{3}{c|}{Office-31} &  \multicolumn{4}{c|}{Office-Home} & \multicolumn{2}{c|}{VisDA-2017} & \multicolumn{3}{c}{Digits} \\
%			\midrule
%			Domains & A & D & W & Ar & Cl & Pr & Rw & Synthetic & Real & S & M & U \\
%			
%			\#Samples & 2,817 & 498 & 795 & 2,427 & 4,365 & 4,439 & 4,357 & 152,397 & 55,388 & 99,289 & 70,000 & 9,298 \\
%			
%			\#Categories & \multicolumn{3}{c|}{31} & \multicolumn{4}{c|}{65} & \multicolumn{2}{c|}{12} & \multicolumn{3}{c}{10} \\
%			\bottomrule
%		\end{tabular}
%	}
%\end{table}	
%
%\begin{figure}[!t]
%	\begin{center}
%		\includegraphics[width=0.8\linewidth]{images/images_show.pdf}
%	\end{center}
%	\caption{Several images sampled from each domain are shown in each column.
%	}
%	\label{fig:images_show}
%\end{figure}

\textbf{Digits} is a commonly used benchmark that contains SVHN (\textbf{S}) \cite{svhn}, MNIST (\textbf{M}) \cite{mnist_lenet}, and USPS (\textbf{U}) \cite{usps}. SVHN has colored images of multiple blurred digits cropped from real scenes. MNIST includes grayscale digit images with a clean background. USPS involves grayscale hand-written digit images with unconstrained writing style. Each digits dataset has a training set and a test set. we follow the evaluation protocol in \cite{adr,mcd} and use the training set for training and the test set for testing. We evaluate on the four tasks of \textbf{S}$\rightarrow$\textbf{M}, \textbf{M}$\rightarrow$\textbf{U}, \textbf{M*}$\rightarrow$\textbf{U*}, and \textbf{U}$\rightarrow$\textbf{M}. For \textbf{M}$\rightarrow$\textbf{U} and \textbf{M*}$\rightarrow$\textbf{U*}, part or all instances of training sets of MNIST and USPS are used for training respectively. 

We provide the statistical details of the domains of the four datasets in the appendix.

\subsection{Implementation Details}
\label{subsec:52}

For Office-31, Office-Home, and VisDA-2017, we follow the standard evaluation protocol for unsupervised domain adaptation \cite{dann,symnets}. For each task, all labeled source instances and all unlabeled target instances are used as training data, and we evaluate different methods on unlabeled target training data. For Office-31 and Office-Home, based on ResNet-50 \cite{resnet}, we report classification results of mean($\pm$standard deviation) on center-crop images over three random trials. For VisDA-2017, based on ResNet-101 \cite{resnet}, we report the classification result of each category. All results are obtained from the target task classifier $F^t(\cdot)$. Each base network is pre-trained on ImageNet \cite{imagenet}. 
We implement our proposed methods by \textbf{PyTorch}. We fine-tune $G(\cdot)$ and train $F(\cdot)$ from scratch via adversarial training where the learning rate of $F(\cdot)$ is $10$ times that of $G(\cdot)$. We follow \cite{dann} to use the SGD optimizer with momentum $0.9$ and weight decay $0.0001$, and the training schedule: the learning rate of $F(\cdot)$ is adjusted by $\eta_p=\frac{\eta_0}{(1+\alpha p)^\beta}$, where $p$ denotes the process of training epochs that is normalized to be in $[0, 1]$, and we set $\eta_0 = 0.01$, $\alpha = 10$, and $\beta = 0.75$; the penalty $\lambda$ is increased from $0$ to $1$ by $\lambda_{p}=\frac{2}{1+\exp(-\gamma p)}-1$, where we set $\gamma = 10$. 
For Digits, we follow \cite{adr,mcd} to use LeNet \cite{mnist_lenet} as the backbone network, adopt the same experimental setting, and report the classification result of mean$\pm$standard deviation over five random trials.

\subsection{Quantitative and Qualitative Validation}
\label{subsec:53}

\renewcommand{\arraystretch}{1.5}
\begin{table}[t]
	\fontsize{6}{6}\selectfont	
	\centering
	\caption{Ablation study (\%) on Office-31 (ResNet-50). Please refer to Section \ref{subsec:53} to know these methods.}
	\label{table:results_office31_ablation_study}
	\begin{threeparttable}
		\begin{tabular}{cccccccc}
			\toprule
			Methods                & A $\rightarrow$ W & D $\rightarrow$ W & W $\rightarrow$ D & A $\rightarrow$ D & D $\rightarrow$ A & W $\rightarrow$ A & Avg \\
			\midrule
			No Adaptation \cite{resnet} & 78.7$\pm$0.1 & 96.4$\pm$0.1 & 99.3$\pm$0.1 & 83.1$\pm$0.2 & 64.7$\pm$0.0 & 65.9$\pm$0.1 & 81.4 \\
			
			DANN \cite{dann}       & 81.7$\pm$0.2 & 98.0$\pm$0.2 & 99.8$\pm$0.0 & 83.9$\pm$0.7 & 66.4$\pm$0.2 & 66.0$\pm$0.3 & 82.6 \\
			
			MADA \cite{mada}       & 90.0$\pm$0.1 & 97.4$\pm$0.1 & 99.6$\pm$0.1 & 87.8$\pm$0.2 & 70.3$\pm$0.3 & 66.4$\pm$0.3 & 85.2 \\
			
			RCA \cite{rca}         & 90.4$\pm$0.2 & 98.8$\pm$0.1 & \textbf{100.0}$\pm$0.0 & 87.6$\pm$0.3 & 72.2$\pm$0.3 & 72.6$\pm$0.2 & 86.9 \\
			
			SymNet \cite{symnets}  & 87.9$\pm$0.1 & 98.4$\pm$0.2 & 99.9$\pm$0.1 & 90.8$\pm$0.5 & 67.4$\pm$0.6 & 69.7$\pm$0.7 & 85.7 \\
			
			ViDANN                 & 82.8$\pm$0.2 & 97.5$\pm$0.2 & 99.8$\pm$0.0 & 84.6$\pm$0.1 & 66.6$\pm$0.1 & 66.1$\pm$0.2 & 82.9 \\
			
			ViRCA                  & 91.4$\pm$0.1 & 99.1$\pm$0.1 & \textbf{100.0}$\pm$0.0 & 89.2$\pm$0.5 &73.0$\pm$0.3 & 73.9$\pm$0.5 & 87.8 \\
			
			\hline
			CatDA (w/o D-adv and C-adv) & 82.8$\pm$0.1 & 98.6$\pm$0.1 & 99.9$\pm$0.1 & 84.3$\pm$0.1 & 66.9$\pm$0.3 & 66.7$\pm$0.0 & 83.2 \\
			
			CatDA (w/o C-adv)   & 85.2$\pm$0.6 & 98.4$\pm$0.1 & 99.9$\pm$0.1 & 85.2$\pm$0.3 & 70.1$\pm$0.3 & 68.1$\pm$0.3 & 84.5 \\
			
			ViCatDA (w/o VC-adv) & 85.0$\pm$0.2 & 98.8$\pm$0.1 & \textbf{100.0}$\pm$0.0 & 85.4$\pm$0.2 & 70.4$\pm$0.1 & 68.3$\pm$0.0 & 84.7 \\
			
			CatDA (w/o D-adv)     & 91.2$\pm$0.1 & 99.1$\pm$0.1 & \textbf{100.0}$\pm$0.0 & 90.2$\pm$0.3 & 73.0$\pm$0.7 & 71.1$\pm$0.3 & 87.4 \\
			
			ViCatDA (w/o VD-adv) & 91.9$\pm$0.1 & \textbf{99.2}$\pm$0.0 & \textbf{100.0}$\pm$0.0 & 91.1$\pm$0.2 & 74.7$\pm$0.3 & 71.2$\pm$0.2 & 88.0 \\
			
			CatDA (w. same-domain weighting) & 85.9$\pm$0.2 & 98.1$\pm$0.2 & \textbf{100.0}$\pm$0.0 & 85.5$\pm$0.2 & 71.1$\pm$0.1 & 69.6$\pm$0.2 & 85.0 \\
			
			CatDA (w. mixup)        & 94.0$\pm$0.4 & 98.8$\pm$0.1 & \textbf{100.0}$\pm$0.0 & 91.1$\pm$0.5 & 75.6$\pm$0.1 & 73.5$\pm$0.3 & 88.8 \\
			
			CatDA                  & 94.5$\pm$0.6 & 99.0$\pm$0.1 & \textbf{100.0}$\pm$0.0 & 90.7$\pm$0.4 & 74.2$\pm$0.2 & 73.2$\pm$0.1 & 88.6 \\
			
			ViCatDA                & 94.5$\pm$0.2 & \textbf{99.2}$\pm$0.1 & \textbf{100.0}$\pm$0.0 & 92.3$\pm$0.1 & 76.5$\pm$0.2 & 74.2$\pm$0.1 & 89.5 \\
			
			ViCatDA+TDSR                & \textbf{94.7}$\pm$0.3 & \textbf{99.2}$\pm$0.0 & \textbf{100.0}$\pm$0.0 & \textbf{94.4}$\pm$0.3 & \textbf{76.8}$\pm$0.1 & \textbf{74.4}$\pm$0.1 & \textbf{89.9} \\
			\bottomrule
		\end{tabular}
	\end{threeparttable}
\end{table}

\paragraph{Ablation Study} We conduct ablation study on Office-31 \cite{office31} based on ResNet-50 \cite{resnet}. We begin with the very baseline ``No Adaptation'', which simply fine-tunes the base network on source data. 
The second, third, fourth, and fifth baselines are respectively DANN \cite{dann}, MADA \cite{mada}, RCA \cite{rca}, and SymNet \cite{symnets}, without applying entropy minimization \cite{em} (for a fair comparison). To test how our VicDA affects DANN and RCA, we apply it to DANN and RCA as the sixth and seventh baselines, denoted by ``ViDANN'' and ``ViRCA'' respectively. To investigate how much the key components of CatDA and ViCatDA improve the performance, we remove both the domain- and category-level adversarial losses of Eq. (\ref{eqn:dom_adv_loss}) and Eq. (\ref{eqn:cat_adv_loss}) and their VicDA versions of Eq. (\ref{eqn:vic_dom_adv_loss}) and Eq. (\ref{eqn:vic_cat_adv_loss}), or only the category-level one of Eq. (\ref{eqn:cat_adv_loss}) and its VicDA version of Eq. (\ref{eqn:vic_cat_adv_loss}), or only the domain-level one of Eq. (\ref{eqn:dom_adv_loss}) and its VicDA version of Eq. (\ref{eqn:vic_dom_adv_loss}) from the overall objectives of Eq. (\ref{eqn:catda}) and Eq. (\ref{eqn:vicatda}), denoted by ``CatDA (w/o D-adv and C-adv)'', ``CatDA (w/o C-adv)'' and ``ViCatDA (w/o VC-adv)'', and ``CatDA (w/o D-adv)'' and ``ViCatDA (w/o VD-adv)'', respectively. To verify the efficacy of our cross-domain weighting scheme, we train a CatDA using pseudo label predictions from the task classifier of one domain to guide category probability predictions of $F(\cdot)$ on this domain, denoted by ``CatDA (w. same-domain weighting)''. To compare VicDA to mixup \cite{mixup}, we train a CatDA with mixup, denoted by ``CatDA (w. mixup)''. 

The results are reported in Table \ref{table:results_office31_ablation_study}. We have the following observations. \textbf{1)} DANN improves over No Adaptation and CatDA (w/o C-adv) improves over CatDA (w/o D-adv and C-adv), certifying the efficacy of the domain-level adversarial loss. \textbf{2)} CatDA (w/o D-adv) outperforms CatDA (w/o D-adv and C-adv) and CatDA (w/o C-adv), testifying the effectiveness of our proposed category-level adversarial loss. \textbf{3)} ViCatDA improves over CatDA (w. mixup) and CatDA, ViCatDA (w/o VD-adv) improves over CatDA (w/o D-adv), ViCatDA (w/o VC-adv) improves over CatDA (w/o C-adv), ViDANN improves over DANN, and ViRCA improves over RCA, verifying the usefulness of VicDA. Note that VicDA cooperates best with CatDA and RCA, which are based on the joint domain-category classifier. \textbf{4)} CatDA significantly outperforms CatDA (w. same-domain weighting), verifying the efficacy of our cross-domain weighting scheme. \textbf{5)} CatDA and ViCatDA exceed MADA, RCA, and SymNet by a large margin, confirming the superiority of our methods on finer category-level domain alignment. The empirical evidence corroborates the method analysis in Section \ref{sec:analysis}. \textbf{6)} ViCatDA+TDSR further improves the performance over ViCatDA, verifying the effectiveness of TDSR on recovering the intrinsic target discrimination.

\begin{figure}[!t]
	\centering
	\subfigure[No Adaptation]{
		\label{fig:t_sne:subfig:a}
		\includegraphics[scale=0.22]{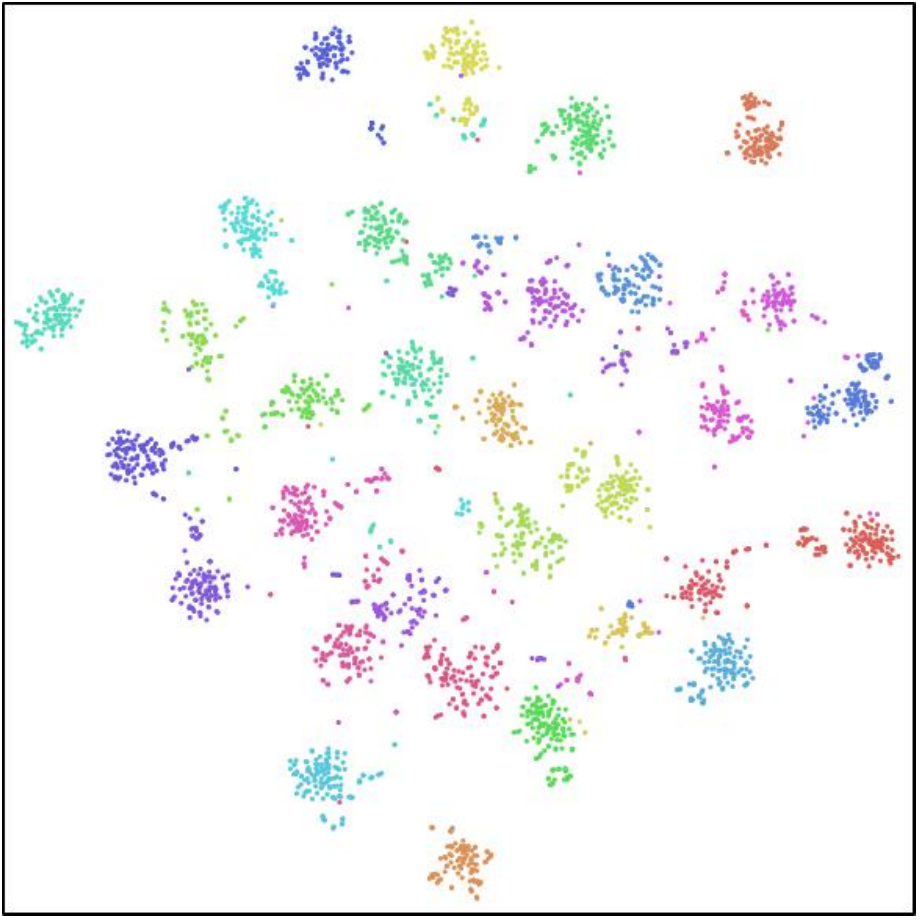}
	}
	\hspace{0.5in}
	\subfigure[DANN]{
		\label{fig:t_sne:subfig:b}
		\includegraphics[scale=0.22]{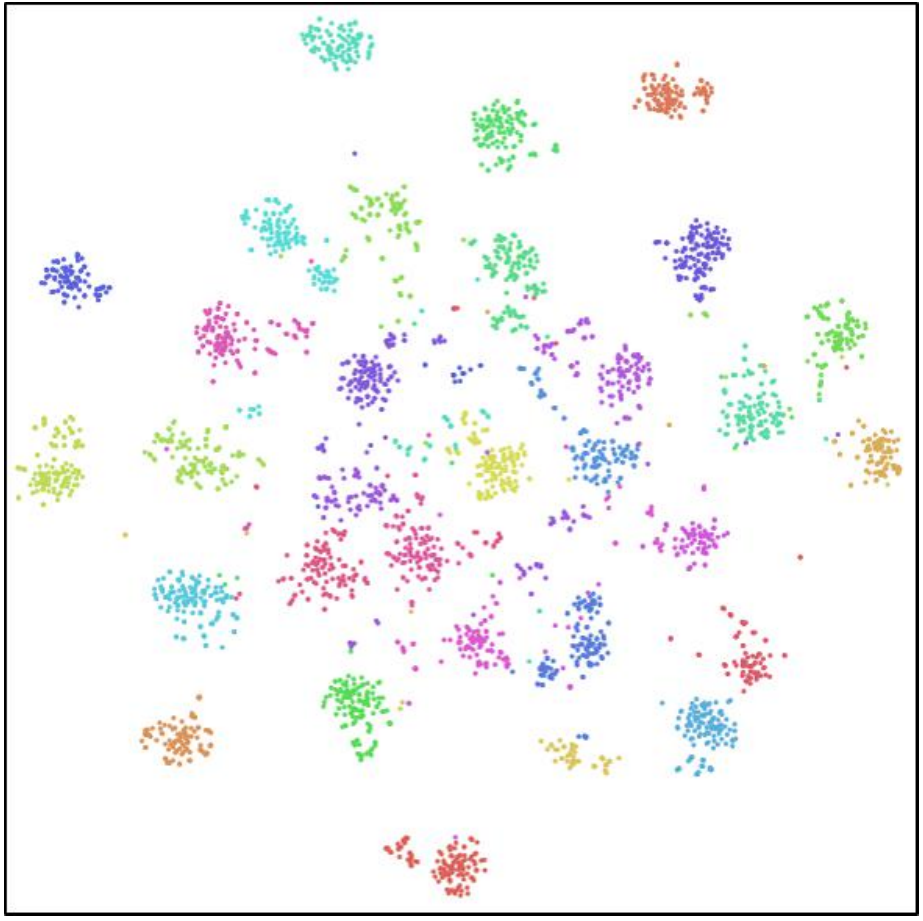}
	}
	\hspace{0.5in}
	\subfigure[CatDA]{
		\label{fig:t_sne:subfig:c}
		\includegraphics[scale=0.22]{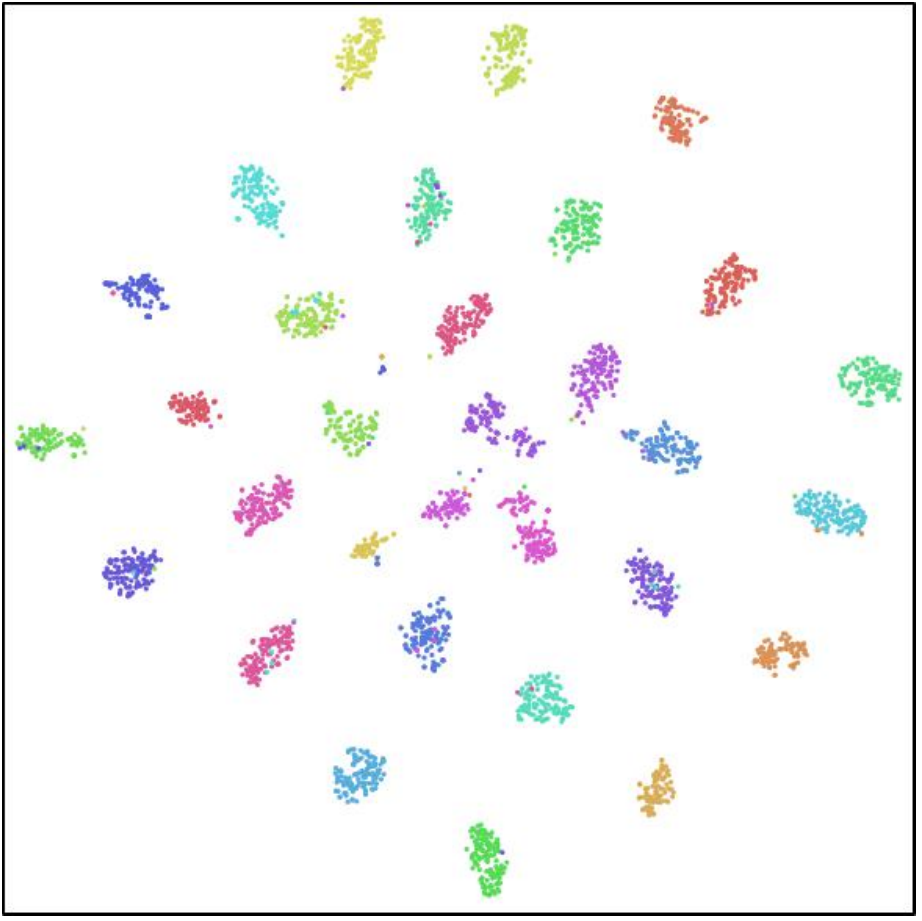}
	}
	\hspace{0.5in}
	\subfigure[ViCatDA]{
		\label{fig:t_sne:subfig:d}
		\includegraphics[scale=0.22]{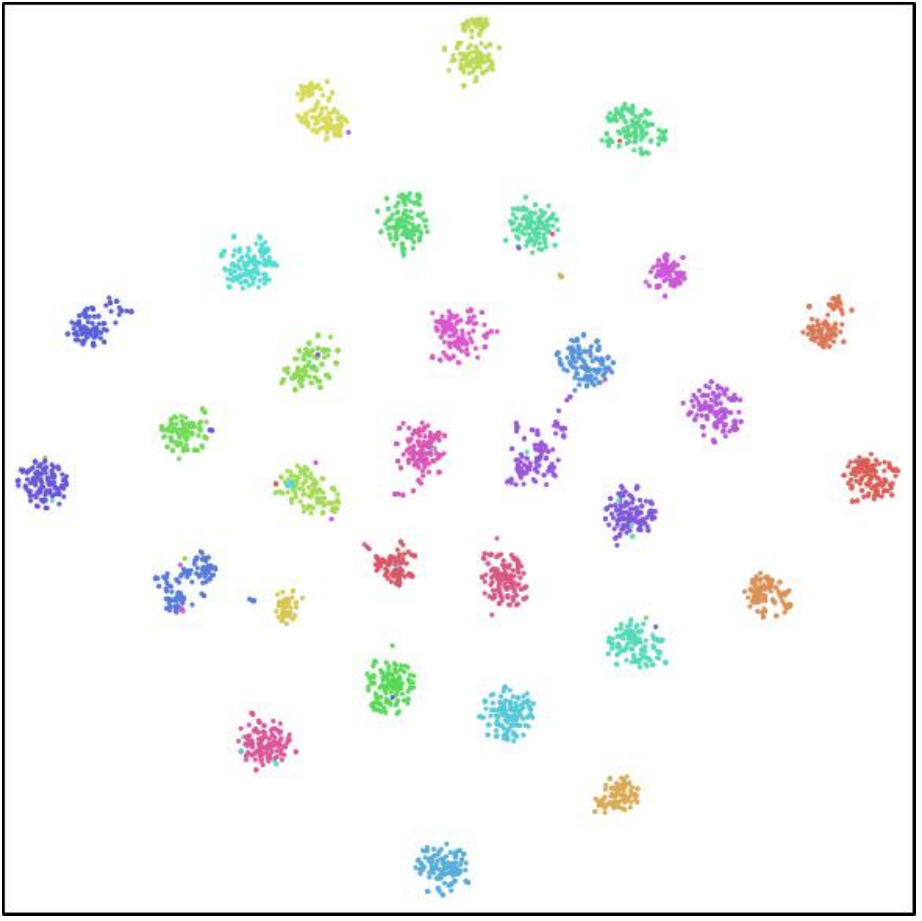}
	}	 
	\caption{ The t-SNE visualization of the feature alignment between the source and target domains. Samples of plotting are from the adaptation task \textbf{A} $\rightarrow$ \textbf{W} in Table \ref{table:results_office31_ablation_study}. Note that different colors denote different categories.}  
	\label{fig:t_sne}
\end{figure}

\paragraph{Feature Visualization} We use t-SNE \cite{t_sne} to visualize features of the source and target domains by No Adaptation, DANN, CatDA, and ViCatDA on \textbf{A} $\rightarrow$ \textbf{W} in Fig. \ref{fig:t_sne}. The two domains are not well aligned by No Adaptation, better aligned by DANN but their corresponding categories are not well aligned. The two domains and their corresponding categories are well aligned while different categories are well discriminated by our methods, confirming their efficacy in achieving the finer category-level alignment.

%\begin{figure}[!t]
%	\begin{center}
%		\includegraphics[scale=0.36]{images/convergence.pdf}
%	\end{center}
%	\caption{Convergence performance by No Adaptation, DANN, CatDA, and ViCatDA on \textbf{A} $\rightarrow$ \textbf{W}. ``($F^s$)'' and ``($F^t$)'' denote results respectively from the source and target task classifiers.
%	}
%	\label{fig:convergence}
%\end{figure}

\begin{figure}[!t]
	\centering
	\subfigure[]{
		\label{fig:convergence}
		\includegraphics[width=0.42\linewidth]{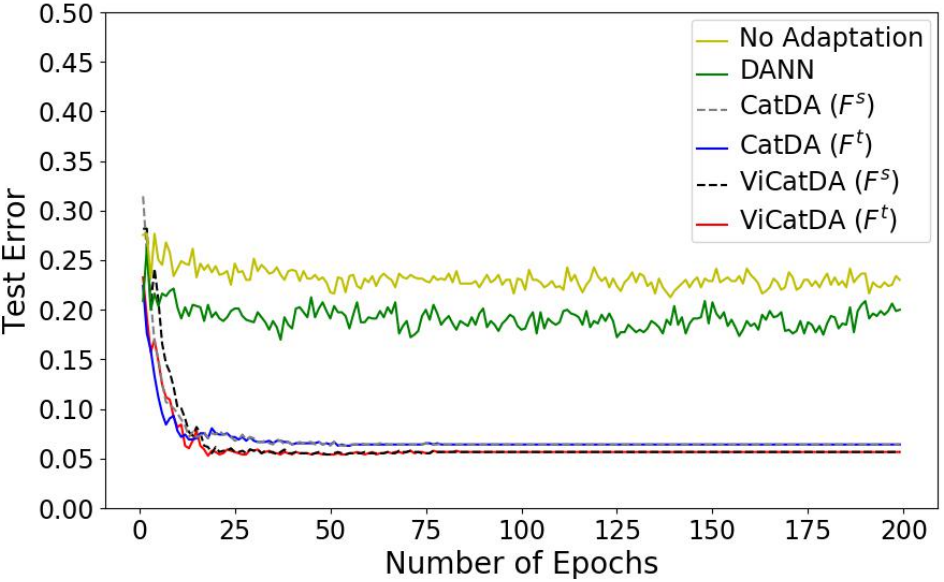}
	}
	\hspace{0.1in}
	\subfigure[]{
		\label{fig:catda_cdw_vs_sdw}
		\includegraphics[width=0.42\linewidth]{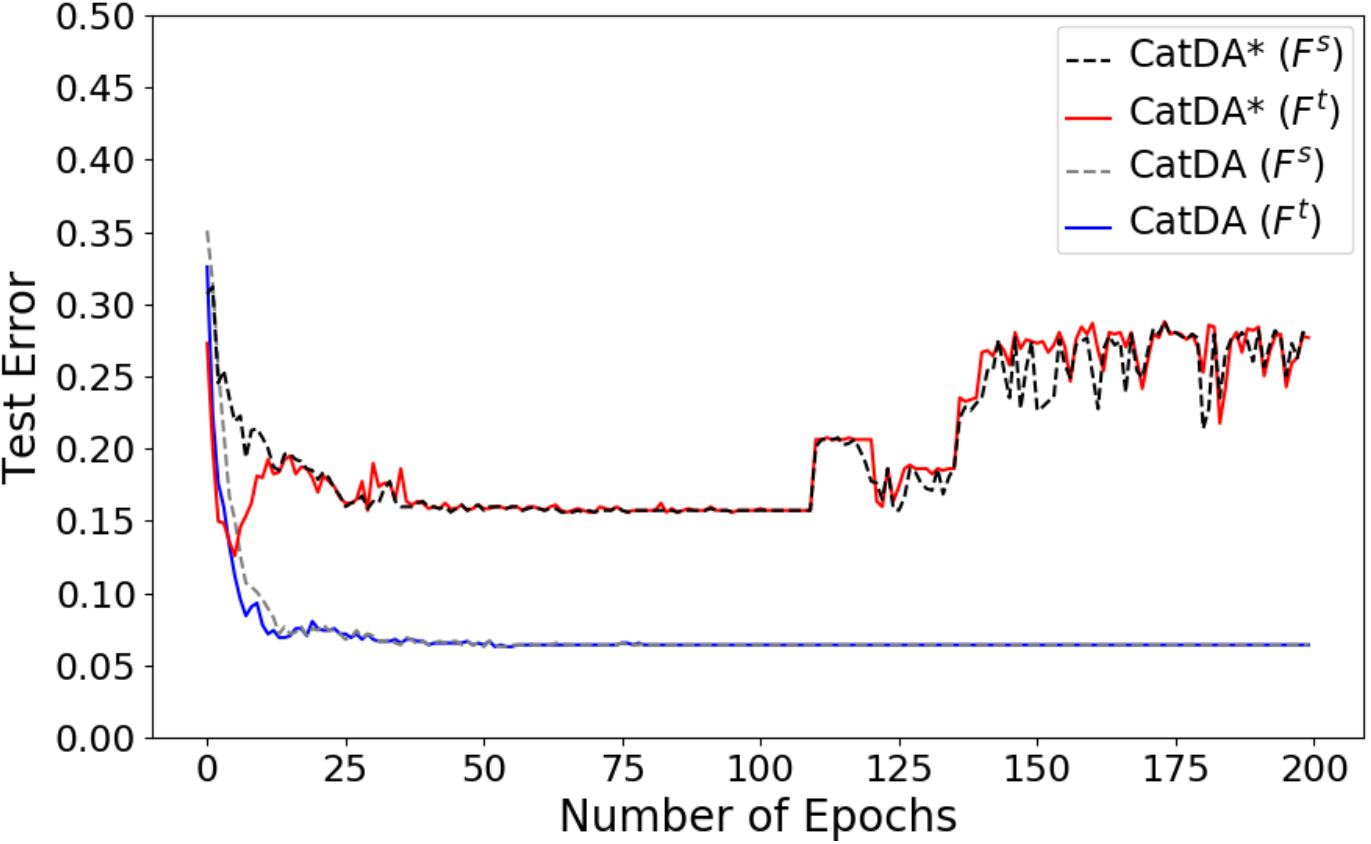}
	}	 
	\caption{(a) Convergence by No Adaptation, DANN, CatDA, and ViCatDA, and (b) training process of our proposed cross-domain weighting scheme (CatDA) and the degenerate same-domain weighting one (CatDA*), on \textbf{A} $\rightarrow$ \textbf{W}. ``($F^s$)'' and ``($F^t$)'' denote results respectively from the source and target task classifiers.
	}  
	\label{fig:convergence_training_process}
\end{figure}

\begin{figure}[!t]
	\centering
	\subfigure[ViCatDA ($F^s$)]{
		\label{fig:conf_matrix:subfig:a}
		\includegraphics[scale=0.29]{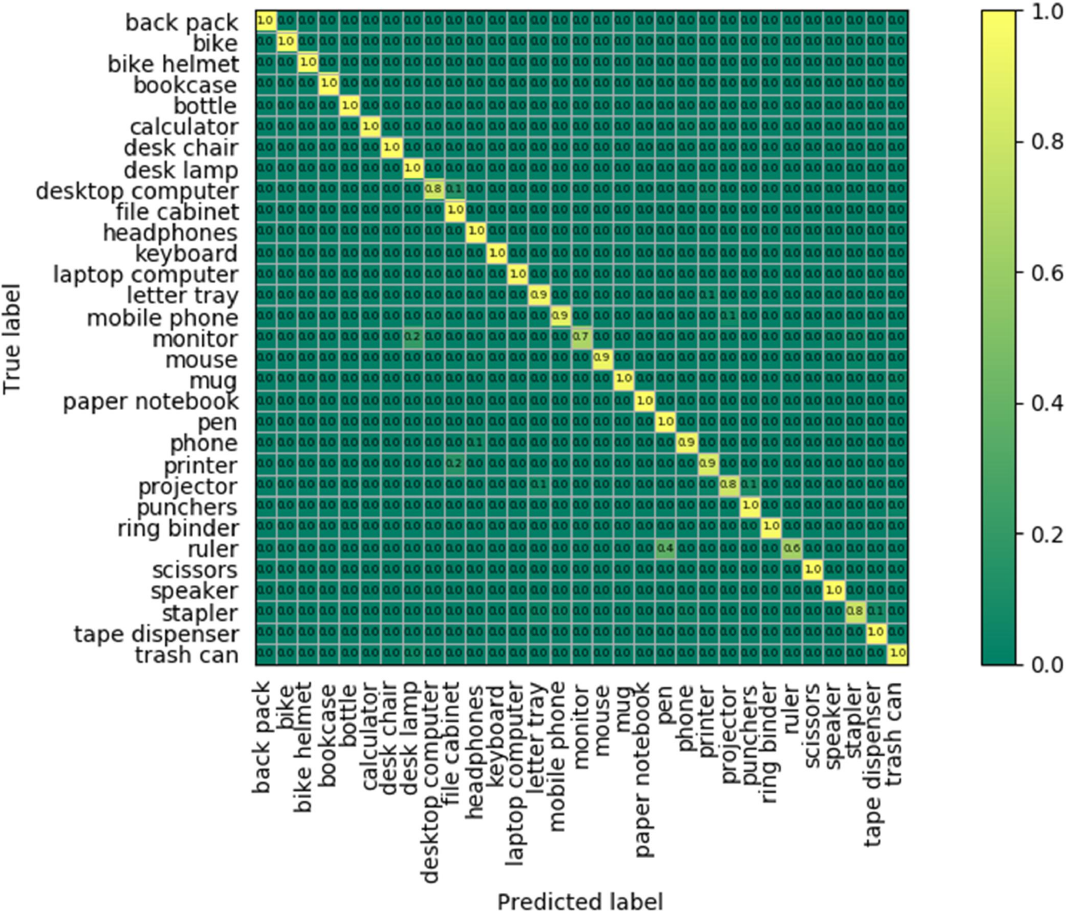}
	}
	\hspace{0.38in}
	\subfigure[ViCatDA ($F^t$)]{
		\label{fig:conf_matrix:subfig:b}
		\includegraphics[scale=0.29]{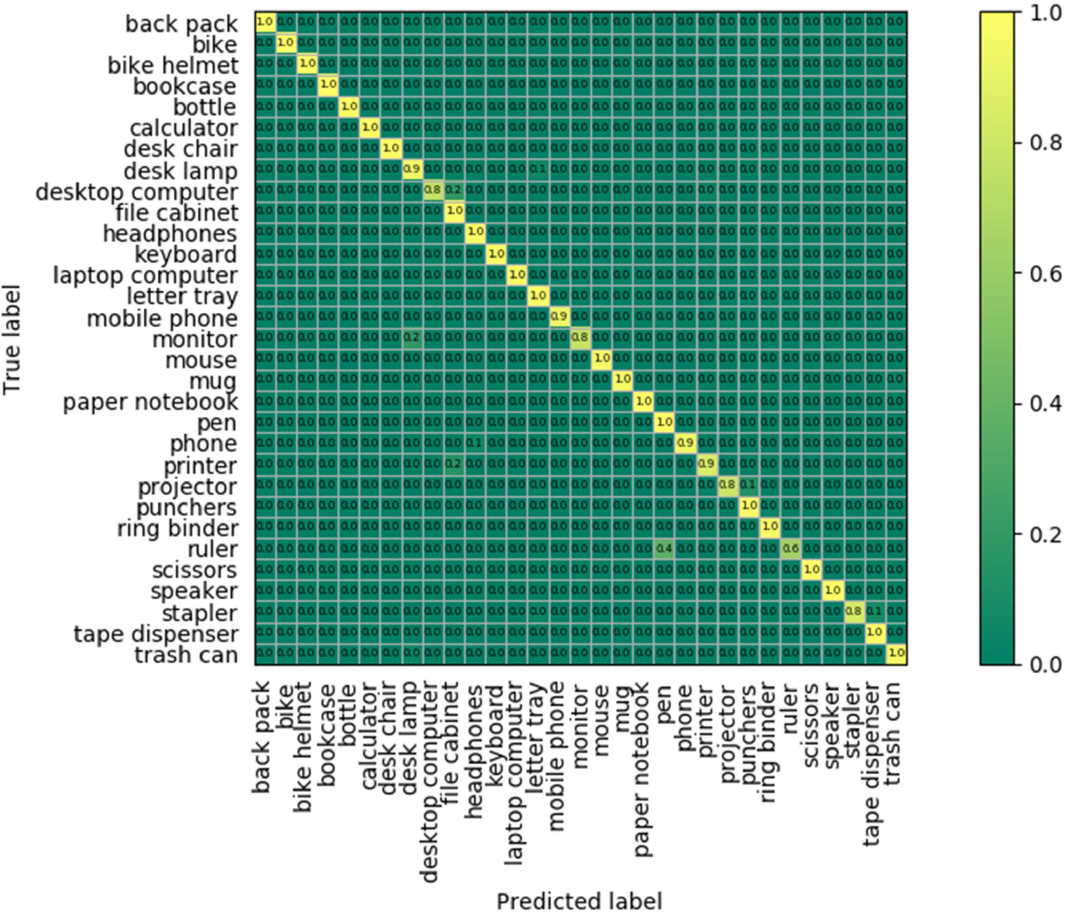}
	}	 
	\caption{Confusion matrix for the target domain on \textbf{A} $\rightarrow$ \textbf{W}. (Zoom in to see the exact class names!)}  
	\label{fig:conf_matrix}
\end{figure}

\paragraph{Convergence Performance} We compare the convergence performance of No Adaptation, DANN, CatDA, and ViCatDA, in terms of test error on \textbf{A} $\rightarrow$ \textbf{W}, in Fig. \ref{fig:convergence}. We can observe that our methods converge faster and smoother than the compared ones. In the early stage of training (e.g. the first $25$ epochs), the test error of CatDA decreases rapidly and then stabilizes at a certain level, indicating the improvement of training reliability. 
Fig. \ref{fig:catda_cdw_vs_sdw} compares the training process of our proposed heterogeneous, cross-domain weighting scheme (CatDA) and the degenerate same-domain weighting one (CatDA*). We can observe that in the early stage of training (e.g. the first $25$ epochs), CatDA has a smaller test error with a smaller fluctuation than CatDA*, indicating that our proposed heterogeneous, cross-domain weighting scheme indeed improves the reliability and stability of model training. Especially, in CatDA, $F^t$ and $F^s$ synchronously improve the classification of target data whereas in CatDA*, the target task classifier $F^t$ degenerates to the source one $F^s$ in terms of the test error. This reflects that our proposed CatDA takes advantage of the complementarity between discriminative information of the source and target domains.

\begin{figure}[!t]
	\begin{center}
		\includegraphics[scale=0.38]{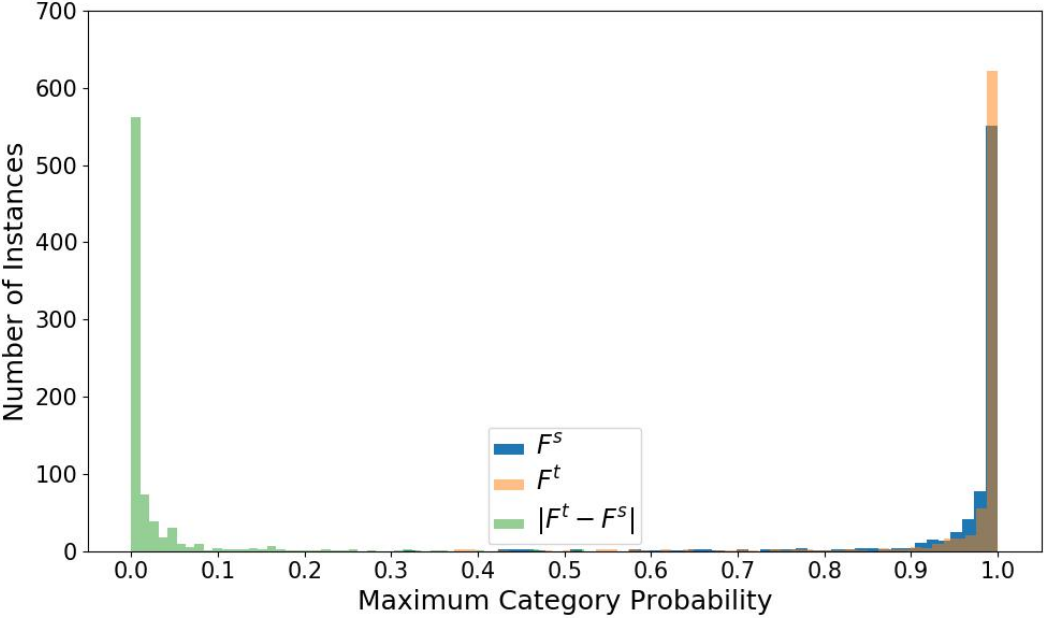}
	\end{center}
	\caption{Statistical histogram of maximum category probabilities predicted by $F^s(\cdot)$ and $F^t(\cdot)$ of ViCatDA, and their prediction discrepancy for all target instances on \textbf{A} $\rightarrow$ \textbf{W}.}
	\label{fig:pred_probs}
\end{figure}

\paragraph{Consistency Verification} In Fig. \ref{fig:conf_matrix}, we report confusion matrices when our ViCatDA converges, i.e. class-wise classification accuracy by two task classifiers of source $F^s(\cdot)$ and target $F^t(\cdot)$ on \textbf{A}$\to$\textbf{W}. We also draw statistical histograms of maximum category probabilities predicted by $F^s(\cdot)$ and $F^t(\cdot)$, and their prediction discrepancy in Fig. \ref{fig:pred_probs}. We can observe the consistency between $F^s(\cdot)$ and $F^t(\cdot)$ in terms of both the predicted category label and prediction confidence, verifying the method analysis in Section \ref{subsec:41}.

\paragraph{Parameter Sensitivity} To evaluate the effect of $\alpha \sim {\rm Beta}(\beta, \beta)$, we do experiments by varying $\beta \in \{0.2, 0.4, 0.6, 0.8, 1.0\}$. Beta distributions with different $\beta$ are illustrated in the appendix. Table \ref{table:results_office31_evaluate_alpha} reports the results on the commonly used Office-31 \cite{office31} benchmark and ResNet-50 \cite{resnet} backbone. We can observe that with a higher probability density about $0.5$, i.e. a larger $\beta$, the averaged classification performance degrades. This suggests that more signals of vicinal domain adaptation would be more beneficial for adaptation on the original source and target domains.

\renewcommand{\arraystretch}{1.5}
\begin{table}[!t]
	\fontsize{6}{6}\selectfont	
	\centering
	\caption{Effect evaluation (\%) of the parameter $\alpha \sim {\rm Beta}(\beta, \beta)$ by varying $\beta$ on Office-31 (ResNet-50).}
	\label{table:results_office31_evaluate_alpha}
	\begin{threeparttable}
		\begin{tabular}{cccccccc}
			\toprule
			Methods                & A $\rightarrow$ W & D $\rightarrow$ W & W $\rightarrow$ D & A $\rightarrow$ D & D $\rightarrow$ A & W $\rightarrow$ A & Avg \\
			\midrule				
			$\beta=0.2$                & \textbf{94.5}$\pm$0.2 & \textbf{99.2}$\pm$0.1 & \textbf{100.0}$\pm$0.0 & \textbf{92.3}$\pm$0.1 & \textbf{76.5}$\pm$0.2 & \textbf{74.2}$\pm$0.1 & \textbf{89.5} \\
			
			$\beta=0.4$                & 94.0$\pm$0.4 & \textbf{99.2}$\pm$0.1 & \textbf{100.0}$\pm$0.0 & 91.4$\pm$0.4 & 75.3$\pm$0.3 & 73.2$\pm$0.4 & 88.9 \\
			
			$\beta=0.6$                & 93.4$\pm$0.9 & \textbf{99.2}$\pm$0.0 & \textbf{100.0}$\pm$0.0 & 90.8$\pm$0.6 & 75.7$\pm$0.8 & 72.2$\pm$0.4 & 88.6 \\
			
			$\beta=0.8$                & 92.2$\pm$0.4 & \textbf{99.2}$\pm$0.1 & \textbf{100.0}$\pm$0.0 & 90.8$\pm$0.8 & 75.4$\pm$0.2 & 71.6$\pm$0.7 & 88.2 \\
			
			$\beta=1.0$                & 90.6$\pm$0.3 & \textbf{99.2}$\pm$0.1 & \textbf{100.0}$\pm$0.0 & 90.6$\pm$0.6 & 74.6$\pm$0.5 & 70.2$\pm$0.4 & 87.5 \\
			\bottomrule
		\end{tabular}
	\end{threeparttable}
\end{table}

\paragraph{Complementation to Popular Techniques} Both entropy minimization (ENT) \cite{em} and consistency enforcing (CON) \cite{consist_enforce} are classical semi-supervised learning techniques, which are popular in domain adaptation community \cite{rca,dirt_t,larger_norm}; to examine whether our proposed ViCatDA can be complementary to the two techniques, we do experiments that combine ViCatDA with ENT or CON on the realistically significant setting \textbf{Synthetic}$\rightarrow$\textbf{Real} of the VisDA-2017 benchmark \cite{visda2017}. Specifically, ENT enforces the task classifier to output a unimodal distribution over category probabilities for target data, such that decision boundaries lie in the low-density region. CON penalizes the inconsistency between category predictions of perturbed copies of the same target instance. Here, we adopt the same data augmentation operations as \cite{dwt_mec} and use the KL-divergence between probability vectors of the two different copies predicted by the joint classifier $F(\cdot)$ as the consistency loss. The results are reported in Table \ref{table:results_visda}. As we can see, with ENT or CON to regularize the target data structure, ViCatDA further improves the classification accuracy of target data by $1.4\%$ and $4.9\%$ respectively.

%In this part, we testify that our ViCatDA can be complementary to other unsupervised domain adaptation techniques to boost the performance. The first technique is entropy minimization (ENT) \cite{em}, which is popular in this community \cite{rca,symnets,dirt_t,larger_norm}. ENT enforces the task classifier to output a unimodal distribution over category probabilities for target data, such that decision boundaries lie in the low-density region. The second technique is consistency enforcing (CON) \cite{consist_enforce}, which penalizes the inconsistency between category predictions of perturbed copies of the same target instance. Here, we adopt the same data augmentation operations as \cite{dwt_mec} and propose to use the KL-divergence between probability vectors of the two different copies predicted by the joint classifier $F(\cdot)$ as the consistency loss. We conduct the case studies on the tough task of \textbf{Synthetic}$\rightarrow$\textbf{Real} and report the results in Table \ref{table:results_visda}. As we can see, with ENT or CON to regularize the target data structure, ViCatDA further improves the classification accuracy of target data by $1.4\%$ and $4.9\%$ respectively.

\renewcommand{\arraystretch}{1.5}
\begin{table}[!t]
	\fontsize{6.5}{6.5}\selectfont	
	\centering
	\caption{Results (\%) on Office-31 (ResNet-50). }
	\label{table:results_office31}
	\begin{threeparttable}		
		\begin{tabular}{cccccccc}
			\toprule
			Methods                & A $\rightarrow$ W & D $\rightarrow$ W & W $\rightarrow$ D & A $\rightarrow$ D & D $\rightarrow$ A & W $\rightarrow$ A & Avg \\
			\midrule
			No Adaptation \cite{resnet} & 78.7$\pm$0.1 & 96.4$\pm$0.1 & 99.3$\pm$0.1 & 83.1$\pm$0.2 & 64.7$\pm$0.0 & 65.9$\pm$0.1 & 81.4 \\
			
			DANN \cite{dann}               & 81.7$\pm$0.2 & 98.0$\pm$0.2 & 99.8$\pm$0.0 & 83.9$\pm$0.7 & 66.4$\pm$0.2 & 66.0$\pm$0.3 & 82.6 \\
			
			%ADDA \cite{adda}               & 86.2$\pm$0.5 & 96.2$\pm$0.3 & 98.4$\pm$0.3 & 77.8$\pm$0.3 & 69.5$\pm$0.4 & 68.9$\pm$0.5 & 82.9 \\
			
			JAN-A \cite{jan}               & 86.0$\pm$0.4 & 96.7$\pm$0.3 & 99.7$\pm$0.1 & 85.1$\pm$0.4 & 69.2$\pm$0.4 & 70.7$\pm$0.5 & 84.6 \\
			
			MADA \cite{mada}               & 90.0$\pm$0.1 & 97.4$\pm$0.1 & 99.6$\pm$0.1 & 87.8$\pm$0.2 & 70.3$\pm$0.3 & 66.4$\pm$0.3 & 85.2 \\
			
			GAACN \cite{gaacn}             & 90.2 & 98.4 & \textbf{100.0} & 90.4 & 67.4 & 67.7 & 85.6 \\
			
			VADA\cite{dirt_t}              & 86.5$\pm$0.5 & 98.2$\pm$0.4 & 99.7$\pm$0.2 & 86.7$\pm$0.4 & 70.1$\pm$0.4 & 70.5$\pm$0.4 & 85.4 \\
			
			%SimNet \cite{SimNet}           & 88.6$\pm$0.5 & 98.2$\pm$0.2 & 99.7$\pm$0.2 & 85.3$\pm$0.3 & 73.4$\pm$0.8 & 71.8$\pm$0.6 & 86.2 \\
			
			%GTA \cite{gen_to_adapt} & 89.5$\pm$0.5 & 97.9$\pm$0.3 & 99.8$\pm$0.4 & 87.7$\pm$0.5 & 72.8$\pm$0.3 & 71.4$\pm$0.4 & 86.5 \\
			
			MCD \cite{mcd}                  & 88.6$\pm$0.2 & 98.5$\pm$0.1 & \textbf{100.0}$\pm$0.0 & 92.2$\pm$0.2 & 69.5$\pm$0.1 & 69.7$\pm$0.3 & 86.5 \\
			
			RCA \cite{rca}         & 90.4$\pm$0.2 & 98.8$\pm$0.1 & \textbf{100.0}$\pm$0.0 & 87.6$\pm$0.3 & 72.2$\pm$0.3 & 72.6$\pm$0.2 & 86.9 \\
			
			SAFN+ENT \cite{larger_norm}     & 90.1$\pm$0.8 & 98.6$\pm$0.2 & 99.8$\pm$0.0 & 90.7$\pm$0.5 & 73.0$\pm$0.2 & 70.2$\pm$0.3 & 87.1 \\
			
			rRevGrad+CAT \cite{cat}   & 94.4$\pm$0.1 & 98.0$\pm$0.2 & \textbf{100.0}$\pm$0.0 & 90.8$\pm$1.8 & 72.2$\pm$0.6 & 70.2$\pm$0.1 & 87.6 \\ %tensorflow
			
			%CDAN+E \cite{cdan}              & 94.1$\pm$0.1 & 98.6$\pm$0.1 & \textbf{100.0}$\pm$0.0 & 92.9$\pm$0.2 & 71.0$\pm$0.3 & 69.3$\pm$0.3 & 87.7 \\
			
			CTSN \cite{ctsn}              & 90.6$\pm$0.3 & 98.6$\pm$0.5 & 99.9$\pm$0.1 & 89.3$\pm$0.3 & 73.7$\pm$0.4 & 74.1$\pm$0.3 & 87.7 \\
			
			SymNet+ENT \cite{symnets}          & 90.8$\pm$0.1 & 98.8$\pm$0.3 & \textbf{100.0}$\pm$0.0 & 93.9$\pm$0.5 & 74.6$\pm$0.6 & 72.5$\pm$0.5 & 88.4 \\
			
			TAT \cite{tat}                  & 92.5$\pm$0.3 & \textbf{99.3}$\pm$0.1 & \textbf{100.0}$\pm$0.0 & 93.2$\pm$0.2 & 73.1$\pm$0.3 & 72.1$\pm$0.3 & 88.4 \\
			
			BSP+CDAN \cite{bsp}             & 93.3$\pm$0.2 & 98.2$\pm$0.2 & \textbf{100.0}$\pm$0.0 & 93.0$\pm$0.2 & 73.6$\pm$0.3 & 72.6$\pm$0.3 & 88.5 \\
			
			\hline
			\textbf{CatDA}                  & 94.5$\pm$0.6 & 99.0$\pm$0.1 & \textbf{100.0}$\pm$0.0 & 90.7$\pm$0.4 & 74.2$\pm$0.2 & 73.2$\pm$0.1 & 88.6 \\
			
			\textbf{ViCatDA}                & 94.5$\pm$0.2 & 99.2$\pm$0.1 & \textbf{100.0}$\pm$0.0 & 92.3$\pm$0.1 & 76.5$\pm$0.2 & 74.2$\pm$0.1 & 89.5 \\
			
			\textbf{ViCatDA+TDSR}                & \textbf{94.7}$\pm$0.3 & 99.2$\pm$0.0 & \textbf{100.0}$\pm$0.0 & \textbf{94.4}$\pm$0.3 & \textbf{76.8}$\pm$0.1 & \textbf{74.4}$\pm$0.1 & \textbf{89.9} \\
			
			\bottomrule
		\end{tabular}
	\end{threeparttable}
\end{table}

%\paragraph{Target Discriminative Structure Recovery} Since adversarial feature alignment may distort the original discriminability of target data \cite{bsp,tat}, we propose to discover the intrinsic discriminative structures of features via established clustering algorithms, and then learn the discovered structures. This model uses CatDA or ViCatDA as initialization and is supervised by the assigned cluster labels. We term this method as \emph{Target Discriminative Structure Recovery (TDSR)}, whose objective is 
%\begin{eqnarray}\label{EqnTDSR}
%\begin{aligned}
%{\cal{L}}_{tdsr} (G, F^t) = -\frac{1}{n_t} \sum_{j=1}^{n_t} \log p_{{\hat{y}}_j^t}^t(\mathbf{x}_j^t), 
%\end{aligned}
%\end{eqnarray}
%where ${\hat{y}}_j^t$ is the attached cluster label of $\mathbf{x}_j^t$. We adopt spherical K-means to cluster target data, which uses $K$ pseudo-labeled target centroids by the learned $F^t(\cdot)$ as initial cluster centers. We conduct experiments on the commonly used Digits and the results are reported in Table \ref{table:results_digits}. With TDSR to recover the intrinsic target discrimination, we can further enhance the performance, demonstrating the usefulness of TDSR.

\begin{table}[!t]
	\centering
	\caption{Results (\%) on Office-Home (ResNet-50). }
	\label{table:results_officehome}
	\resizebox{0.87\textwidth}{!}{	
		\begin{tabular}{cccccccccccccc}
			\toprule
			Methods                         & Ar$\rightarrow$Cl & Ar$\rightarrow$Pr & Ar$\rightarrow$Rw & Cl$\rightarrow$Ar & Cl$\rightarrow$Pr & Cl$\rightarrow$Rw & Pr$\rightarrow$Ar & Pr$\rightarrow$Cl & Pr$\rightarrow$Rw & Rw$\rightarrow$Ar & Rw$\rightarrow$Cl & Rw$\rightarrow$Pr & Avg  \\
			\midrule
			No Adaptation \cite{resnet}     & 34.9      & 50.0     & 58.0      & 37.4      & 41.9      & 46.2     & 38.5     & 31.2     & 60.4     & 53.9     & 41.2     & 59.9 & 46.1 \\
			
			DAN \cite{dan}                  & 43.6     & 57.0     & 67.9      & 45.8      & 56.5      & 60.4     & 44.0     & 43.6     & 67.7     & 63.1     & 51.5     & 74.3  & 56.3 \\
			
			DANN \cite{dann}                & 45.6     & 59.3     & 70.1      & 47.0      & 58.5      & 60.9     & 46.1     & 43.7     & 68.5     & 63.2     & 51.8      & 76.8 & 57.6 \\
			
			JAN \cite{jan}                  & 45.9     & 61.2     & 68.9      & 50.4      & 59.7      & 61.0     & 45.8     & 43.4     & 70.3     & 63.9     & 52.4      & 76.8 & 58.3 \\
			
			SE \cite{selfensembling}        & 48.8 & 61.8 & 72.8 & 54.1 & 63.2 & 65.1 & 50.6 & 49.2 & 72.3 & 66.1 & 55.9 & 78.7 & 61.5 \\
			
			DWT-MEC \cite{dwt_mec}          & 50.3 & 72.1 & 77.0 & 59.6 & 69.3 & 70.2 & 58.3 & 48.1 & 77.3 & 69.3 & 53.6 & 82.0 & 65.6 \\
			
			%CDAN+E \cite{cdan}  & 50.7      & 70.6     & 76.0     & 57.6       & 70.0      & 70.0     & 57.4     & 50.9     & 77.3      & 70.9      & 56.7     & 81.6 & 65.8 \\
			
			TAT \cite{tat}                  & 51.6 & 69.5 & 75.4 & 59.4 & 69.5 & 68.6 & 59.5 & 50.5 & 76.8 & 70.9 & 56.6 & 81.6 & 65.8 \\
			
			GAACN \cite{gaacn}              & 53.1 & 71.5 & 74.6 & 59.9 & 64.6 & 67.0 & 59.2 & 53.8 & 75.1 & 70.1 & 59.3 & 80.9 & 65.8 \\
			
			BSP+CDAN \cite{bsp}             & 52.0 & 68.6 & 76.1 & 58.0 & 70.3 & 70.2 & 58.6 & 50.2 & 77.6 & 72.2 & 59.3 & 81.9 & 66.3 \\
			
			SAFN \cite{larger_norm}         & 52.0 & 71.7 & 76.3 & 64.2 & 69.9 & 71.9 & 63.7 & 51.4 & 77.1 & 70.9 & 57.1 & 81.5 & 67.3 \\
			
			SymNet+ENT \cite{symnets}       & 47.7 & 72.9 & 78.5 & 64.2 & 71.3 & 74.2 & 64.2 & 48.8 & 79.5 & \textbf{74.5} & 52.6 & 82.7 & 67.6 \\
			
			\hline
			\textbf{CatDA}                  & 49.3 & 72.8 & 78.2 & 63.7 & 70.7 & 72.5 & 64.3 & 50.2 & 79.2 & 73.4 & 56.7 & 82.3 & 67.8 \\
			
			\textbf{ViCatDA}                & 50.9 & 74.7 & \textbf{78.8} & 64.8 & 71.7 & \textbf{74.4} & \textbf{64.5} & 52.4 & \textbf{80.4} & \textbf{74.5} & 57.4 & \textbf{83.2} & 69.0 \\
			
			\textbf{ViCatDA+TDSR}                & \textbf{56.1} & \textbf{75.4} & \textbf{78.8} & \textbf{65.0} & \textbf{71.9} & \textbf{74.4} & \textbf{64.5} & \textbf{55.1} & \textbf{80.4} & \textbf{74.5} & \textbf{61.1} & \textbf{83.2} & \textbf{70.0} \\
			
			\bottomrule
		\end{tabular}
	}
\end{table}

\begin{table}[!t]
	\centering
	\caption{Results (\%) on VisDA-2017 (ResNet-101). Please refer to Section \ref{subsec:53} to know ENT and CON.}
	\label{table:results_visda}
	\resizebox{0.85\textwidth}{!}{	
		\begin{tabular}{cccccccccccccc}
			\toprule
			Methods                & plane & bcycl & bus & car & horse & knife & mcycl & person & plant & sktbrd & train & truck & mean \\
			\midrule
			No Adaptation \cite{resnet}           & 55.1 & 53.3 & 61.9 & 59.1 & 80.6 & 17.9 & 79.7 & 31.2 & 81.0 & 26.5 & 73.5 & 8.5 & 52.4 \\
			
			DANN \cite{dann}  & 81.9 & 77.7 & 82.8 & 44.3 & 81.2 & 29.5 & 65.1 & 28.6 & 51.9 & 54.6 & 82.8 & 7.8 & 57.4 \\
			
			DAN \cite{dan}         & 87.1 & 63.0 & 76.5 & 42.0 & 90.3 & 42.9 & 85.9 & 53.1 & 49.7 & 36.3 & 85.8 & 20.7 & 61.1 \\
			
			MCD \cite{mcd} & 87.0 & 60.9 & 83.7 & 64.0 & 88.9 & 79.6 & 84.7 & 76.9 & 88.6 & 40.3 & 83.0 & 25.8 & 71.9 \\
			
			GPDA \cite{gpda}  & 83.0 & 74.3 & 80.4 & 66.0 & 87.6 & 75.3 & 83.8 & 73.1 & 90.1 & 57.3 & 80.2 & 37.9 & 73.3 \\
			
			ADR \cite{adr}    & 87.8 & 79.5 & 83.7 & 65.3 & 92.3 & 61.8 & 88.9 & 73.2 & 87.8 & 60.0 & 85.5 & 32.3 & 74.8 \\
			
			BSP+CDAN \cite{bsp} & 92.4 & 61.0 & 81.0 & 57.5 & 89.0 & 80.6 & 90.1 & 77.0 & 84.2 & 77.9 & 82.1 & 38.4 & 75.9 \\
			
			TPN \cite{tpn}    & 93.7 & \textbf{85.1} & 69.2 & \textbf{81.6} & \textbf{93.5} & 61.9 & 89.3 & \textbf{81.4} & 93.5 & 81.6 & 84.5 & \textbf{49.9} & 80.4 \\
			
			\hline
			\textbf{CatDA} & 93.9 & 71.3 & 75.9 & 56.0 & 86.3 & \textbf{92.4} & 86.0 & 80.0 & 87.9 & 55.8 & 89.3 & 40.8 & 76.3 \\
			
			\textbf{ViCatDA} & 93.9 & 67.3 & 78.6 & 66.9 & 89.3 & 88.4 & 91.0 & 77.9 & 90.2 & 68.2 & 88.4 & 31.8 & 77.7 \\
			
			\textbf{ViCatDA+TDSR} & 92.8 & 76.4 & 80.2 & 64.0 & 88.4 & 92.1 & 87.9 & 78.9 & 88.0 & 81.8 & 89.6 & 42.1 & 80.2 \\
			
			\hline
			
			\textbf{ViCatDA+ENT} & 92.2 & 76.4 & 79.3 & 68.1 & 92.2 & 91.5 & 90.4 & 79.8 & \textbf{93.7} & 67.0 & \textbf{90.7} & 28.4 & 79.1 \\
			
			\textbf{ViCatDA+CON} & \textbf{95.9} & 76.5 & \textbf{89.0} & 71.1 & 91.8 & 89.2 & \textbf{92.4} & 79.6 & 92.9 & \textbf{90.8} & 88.8 & 33.3 & \textbf{82.6} \\
			
			\bottomrule
		\end{tabular}
	}
\end{table}

\renewcommand{\arraystretch}{1.5}
\begin{table}[!t]
	\fontsize{6.0}{6.0}\selectfont	
	\centering
	\caption{Results (\%) on Digits (LeNet). }
	\label{table:results_digits}
	\begin{threeparttable}	
		\begin{tabular}{cccccc}
			\hline
			Methods                 & S $\rightarrow$ M & M $\rightarrow$ U & M* $\rightarrow$ U* & U $\rightarrow$ M & Avg \\
			\hline
			No Adaptation \cite{mnist_lenet} & 67.1 & 76.7 & 79.4 & 63.4 & 71.7 \\
			
			DAN \cite{dan}          & 71.1 & - & 81.1 & - & - \\
			
			DANN \cite{dann}        & 71.1 & 77.1$\pm$1.8 & 85.1 & 73.0$\pm$0.2 & 76.6 \\
			
			%ADDA \cite{adda}        & 76.0$\pm$1.8 & 89.4$\pm$0.2 & - & 90.1$\pm$0.8 & - \\
			
			%CoGAN \cite{cogan}      & - & 91.2$\pm$0.8 & - & 89.1$\pm$0.8 & - \\
			
			%DRCN \cite{drcn}        & 82.0$\pm$0.1 & 91.8$\pm$0.09 & - & 73.7$\pm$0.04 & - \\
			
			%DSN \cite{dsns}          & 82.7 & 91.3 & - & - & - \\
			
			%LDC \cite{layerWiseCorrection} & 89.5$\pm$2.1 & - & - & - & - \\
			
			%CyCADA \cite{cycada}    & 90.4$\pm$0.4 & 95.6$\pm$0.2 & - & 96.5$\pm$0.1 & - \\
			
			MSTN \cite{mstn}        & 91.7$\pm$1.5 & 92.9$\pm$1.1 & - & - & - \\
			
			TPN \cite{tpn}          & 93.0 & 92.1 & - & 94.1 & - \\
			
			PFAN \cite{pfan}        & 93.9$\pm$0.8 & 95.0$\pm$1.3 & - & - & - \\
			
			ADR \cite{adr}          & 94.1$\pm$1.37 & 91.3$\pm$0.65 & - & 91.5$\pm$3.61 & - \\
						
			GAACN \cite{gaacn}      & 94.6 & 95.4 & - & \textbf{98.3} & - \\
						
			DM-ADA \cite{dm_ada}    & 95.5$\pm$1.1 & 94.8$\pm$0.7 & 96.7$\pm$0.5 & 94.2$\pm$0.9 & 95.3 \\
			
			ASSC \cite{associativeDA}        & 95.7$\pm$1.5 & - & - & - & - \\
			
			MCD \cite{mcd}          & 96.2$\pm$0.4 & 94.2$\pm$0.7 & 96.5$\pm$0.3 & 94.1$\pm$0.3 & 95.3 \\
			
			CTSN \cite{ctsn}        & 97.1$\pm$0.3 & 96.1$\pm$0.3 & - & 97.3$\pm$0.2 & - \\ 
			
			\hline
			\textbf{CatDA}             & 96.4$\pm$0.2 & 95.0$\pm$0.4 & 97.0$\pm$0.2 & 96.0$\pm$0.2 & 96.1 \\
			\textbf{ViCatDA}           & 97.1$\pm$0.1 & 96.0$\pm$0.2 & 97.9$\pm$0.1 & 96.7$\pm$0.2 & 96.9 \\
			\textbf{ViCatDA+TDSR}      & \textbf{97.2}$\pm$0.1 & \textbf{96.7}$\pm$0.1 & \textbf{98.1}$\pm$0.1 & 97.1$\pm$0.1 & \textbf{97.3} \\
			\hline
		\end{tabular}
	\end{threeparttable}	
\end{table}

\subsection{Object Classification}
\label{subsec:54}

In this section, we compare our proposed method with existing ones on Office-31 \cite{office31}, Office-Home \cite{officehome}, and VisDA-2017 \cite{visda2017} in Tables \ref{table:results_office31}, \ref{table:results_officehome}, and \ref{table:results_visda} respectively, where results of existing methods are quoted from their respective papers or \cite{cdan,tat,dwt_mec,mcd}. Our proposed ViCatDA improves the performance on hard adaptation tasks, e.g. \textbf{D} $\rightarrow$ \textbf{A} and \textbf{W} $\rightarrow$ \textbf{A}, and on the difficult dataset of Office-Home with more categories and larger size, verifying the effectiveness of ViCatDA. On the realistically significant \textbf{Synthetic}$\rightarrow$\textbf{Real} task, ViCatDA with the KL-divergence based consistency enforcing loss (CON), achieves the best result, confirming the validity of ViCatDA and the excellent effect of consistency enforcing on reducing simulation-to-real shift. 
ViCatDA consistently performs better than the compared methods on the three datasets, testifying its superiority in category-level domain alignment. With the proposed TDSR to recover the intrinsic target discrimination, ViCatDA+TDSR achieves a remarkable performance gain over ViCatDA, demonstrating the necessity and usefulness of TDSR.

\subsection{Digit Classification}
\label{subsec:55}

We show the comparison of different methods on Digits in Table \ref{table:results_digits}. Results of existing methods are quoted from their respective papers or \cite{adr,mcd}. From Table \ref{table:results_digits}, we take several interesting observations. 
\textbf{1)} No Adaptation performs worst, suggesting the existence of domain discrepancy and the necessity of domain adaptation. 
\textbf{2)} Methods based on domain-level domain alignment, e.g. DAN \cite{dan} and DANN \cite{dann}, perform better than No Adaptation, showing their efficacy in learning domain-invariant features. 
\textbf{3)} Methods based on both domain- and category-level domain alignments, e.g. MSTN \cite{mstn} and our ViCatDA, significantly outperform those based on only domain-level domain alignment, which is unaware of classification boundaries and thus causes false alignment between different classes across domains, i.e. negative transfer. It attests that the category-level domain alignment, which exploits the target-discriminative information from the task classifier(s), is essential. 
\textbf{4)} Our ViCatDA (with TDSR) exceeds all compared methods on almost all tasks, verifying its efficacy in reducing the domain gap. 
\textbf{5)} We consistently achieve the new state of the art whether the network is small like LeNet or large like ResNet-101, demonstrating the robustness of our proposed methods.

\section{Conclusion and Future Work}
\label{sec:conclusion}

In this work, based on the joint domain-category classifier, we propose novel losses of adversarial training at multiple levels to promote \emph{categorical domain adaptation (CatDA)}, where the category-level adversarial training improves over the domain-level one by a heterogenous, cross-domain weighting design that enhances the consistency of category predictions between the source and target task classifiers. 
To improve, we generate a (theoretically) infinite number of vicinal domains whose instances are constructed by a convex combination of pairs of instances respectively from the two domains, and propose novel adversarial losses for \emph{vicinal domain adaptation (VicDA)} based on CatDA, leading to our \emph{Vicinal and Categorical Domain Adaptation (ViCatDA)}. 
To recover the intrinsic target discrimination damaged by adversarial feature alignment, we propose \emph{Target Discriminative Structure Recovery (TDSR)} based on semantically anchored spherical k-means. 
We also analyze the working mechanisms of our key designs in principle. 
We achieve the new state of the art on four benchmark datasets.

In future work, we will further improve domain adaptation from three perspectives. 
On the aspect of model, we will design a fine-grained joint classifier, which learns a distribution over not only domain and category but also cluster. Generally, each category has various objects with different appearances or shapes, and its images are taken from diverse viewpoints. According to these variation factors, images of each category can be divided into multiple clusters. %Domain adaptation at the finer cluster level rather than only domain and category levels can further alleviate the false alignment between different categories across domains, i.e. reducing negative transfer. 
On the aspect of algorithm, %for enhancing positive transfer, 
we will explore a more effective and efficient way to generate and align vicinal domains, and conduct more careful studies in different clustering frameworks that discover and utilize the intrinsic target discrimination. 
On the aspect of dataset, to approach practical application, we will collect a large-scale synthetic-to-real dataset with more real-world categories.

\section*{Acknowledgment}
	This work was supported in part by the National Natural Science Foundation of China (Grant No.: 61771201), the Program for Guangdong Introducing Innovative and Enterpreneurial Teams (Grant No.: 2017ZT07X183), and the Guangdong R\&D key project of China (Grant No.: 2019B010155001).

%\section*{References}

    \begin{small}
    	\bibliographystyle{elsarticle-num}
    	\bibliography{references}

\begin{thebibliography}{10}
\expandafter\ifx\csname url\endcsname\relax
  \def\url#1{\texttt{#1}}\fi
\expandafter\ifx\csname urlprefix\endcsname\relax\def\urlprefix{URL }\fi
\expandafter\ifx\csname href\endcsname\relax
  \def\href#1#2{#2} \def\path#1{#1}\fi

\bibitem{resnet}
K.~{He}, X.~{Zhang}, S.~{Ren}, J.~{Sun}, Deep residual learning for image
  recognition, in: Proc. IEEE Conf. Comput. Vis. Pattern Recognit., 2016, pp.
  770--778.

\bibitem{mnist_lenet}
Y.~LeCun, L.~Bottou, Y.~Bengio, P.~Haffner, Gradient-based learning applied to
  document recognition, in: Proceedings of the IEEE, Vol.~86, 1998, p.
  2278–2324.

\bibitem{imagenet}
J.~Deng, W.~Dong, R.~Socher, L.-J. Li, K.~Li, L.~Fei-Fei, Imagenet: A
  large-scale hierarchical image database, in: Proc. IEEE Conf. Comput. Vis.
  Pattern Recognit., 2009, pp. 248--255.

\bibitem{fcn_ss}
J.~Long, E.~Shelhamer, T.~Darrell, Fully convolutional networks for semantic
  segmentation, in: Proc. IEEE Conf. Comput. Vis. Pattern Recognit., 2015, pp.
  3431--3440.

\bibitem{tl_survey}
S.~J. Pan, Q.~Yang, A survey on transfer learning, IEEE Transactions on
  Knowledge and Data Engineering 22 (2010) 1345--1359.

\bibitem{da_theory2}
S.~Ben-David, J.~Blitzer, K.~Crammer, A.~Kulesza, F.~Pereira, J.~W. Vaughan, A
  theory of learning from different domains, Machine Learning 79 (2010)
  151--175.

\bibitem{da_theory1}
S.~Ben-David, J.~Blitzer, K.~Crammer, F.~Pereira, Analysis of representations
  for domain adaptation, in: Proc. Neur. Info. Proc. Sys., 2007, pp. 137--144.

\bibitem{dann}
Y.~Ganin, E.~Ustinova, H.~Ajakan, P.~Germain, H.~Larochelle, F.~Laviolette,
  M.~Marchand, V.~Lempitsky, Domain-adversarial training of neural networks, J.
  Mach. Learn. Res. 17 (2016) 2096--2030.

\bibitem{mada}
Z.~Pei, Z.~Cao, M.~Long, J.~Wang, Multi-adversarial domain adaptation, in:
  Association for the Advancement of Artificial Intelligence, 2018, pp.
  3934--3941.

\bibitem{cdan}
M.~Long, Z.~Cao, J.~Wang, M.~I. Jordan, Conditional adversarial domain
  adaptation, in: Proc. Neur. Info. Proc. Sys., 2018, pp. 1647--1657.

\bibitem{bsp}
X.~Chen, S.~Wang, M.~Long, J.~Wang, Transferability vs. discriminability: Batch
  spectral penalization for adversarial domain adaptation, in: Proc. Int. Conf.
  Mach. Learn., Vol.~97, 2019, pp. 1081--1090.

\bibitem{mstn}
S.~Xie, Z.~Zheng, L.~Chen, C.~Chen, Learning semantic representations for
  unsupervised domain adaptation, in: Proc. Int. Conf. Mach. Learn., Vol.~80,
  2018, pp. 5423--5432.

\bibitem{gans}
I.~Goodfellow, J.~Pouget-Abadie, M.~Mirza, B.~Xu, D.~Warde-Farley, S.~Ozair,
  A.~Courville, Y.~Bengio, Generative adversarial nets, in: Proc. Neur. Info.
  Proc. Sys., 2014, pp. 2672--2680.

\bibitem{rca}
S.~{Cicek}, S.~{Soatto}, Unsupervised domain adaptation via regularized
  conditional alignment, in: Proc. IEEE Int. Conf. Comput. Vis., 2019, pp.
  1416--1425.

\bibitem{symnets}
Y.~{Zhang}, H.~{Tang}, K.~{Jia}, M.~{Tan}, Domain-symmetric networks for
  adversarial domain adaptation, in: Proc. IEEE Conf. Comput. Vis. Pattern
  Recognit., 2019, pp. 5026--5035.

\bibitem{gfk}
B.~{Gong}, Y.~{Shi}, F.~{Sha}, K.~{Grauman}, Geodesic flow kernel for
  unsupervised domain adaptation, in: Proc. IEEE Conf. Comput. Vis. Pattern
  Recognit., 2012.

\bibitem{dlow}
R.~Gong, W.~Li, Y.~Chen, L.~V. Gool, Dlow: Domain flow for adaptation and
  generalization, in: Proc. IEEE Conf. Comput. Vis. Pattern Recognit., 2019,
  pp. 2472--2481.

\bibitem{mixup}
H.~Zhang, M.~Cisse, Y.~N. Dauphin, D.~Lopez-Paz, mixup: Beyond empirical risk
  minimization, in: International Conference on Learning Representations, 2018.

\bibitem{tat}
H.~Liu, M.~Long, J.~Wang, M.~Jordan, Transferable adversarial training: A
  general approach to adapting deep classifiers, in: Proc. Int. Conf. Mach.
  Learn., Vol.~97, 2019, pp. 4013--4022.

\bibitem{da_theory3}
H.~Zhao, R.~T.~D. Combes, K.~Zhang, G.~Gordon, On learning invariant
  representations for domain adaptation, in: Proc. Int. Conf. Mach. Learn.,
  Vol.~97, 2019, pp. 7523--7532.

\bibitem{spherical_k_means}
K.~Hornik, I.~Feinerer, M.~Kober, C.~Buchta, Spherical k-means clustering,
  Journal of Statistical Software 50 (2012) 1--22.

\bibitem{GenEqu}
S.~Arora, R.~Ge, Y.~Liang, T.~Ma, Y.~Zhang, Generalization and equilibrium in
  generative adversarial nets ({GAN}s), in: Proc. Int. Conf. Mach. Learn.,
  2017, pp. 224--232.

\bibitem{GLG_HeUDA}
F.~{Liu}, G.~{Zhang}, J.~{Lu}, Heterogeneous domain adaptation: An unsupervised
  approach, IEEE Transactions on Neural Networks and Learning Systems 31 (2020)
  5588--5602.

\bibitem{adda}
E.~{Tzeng}, J.~{Hoffman}, K.~{Saenko}, T.~{Darrell}, Adversarial discriminative
  domain adaptation, in: Proc. IEEE Conf. Comput. Vis. Pattern Recognit., 2017,
  pp. 2962--2971.

\bibitem{BeyondSW}
A.~Rozantsev, M.~Salzmann, P.~Fua, Beyond sharing weights for deep domain
  adaptation, IEEE Trans. Pattern Anal. Mach. Intell. 41 (2019) 801--814.

\bibitem{cmd}
W.~Zellinger, T.~Grubinger, E.~Lughofer, T.~Natschl{\"a}ger, S.~Saminger-Platz,
  Central moment discrepancy (cmd) for domain-invariant representation
  learning, in: Proc. Int. Conf. on Learn. Rep., 2017.

\bibitem{SimNet}
P.~O. {Pinheiro}, Unsupervised domain adaptation with similarity learning, in:
  Proc. IEEE Conf. Comput. Vis. Pattern Recognit., 2018, pp. 8004--8013.

\bibitem{gen_to_adapt}
S.~{Sankaranarayanan}, Y.~{Balaji}, C.~D. {Castillo}, R.~{Chellappa}, Generate
  to adapt: Aligning domains using generative adversarial networks, in: Proc.
  IEEE Conf. Comput. Vis. Pattern Recognit., 2018, pp. 8503--8512.

\bibitem{dwt_mec}
S.~{Roy}, A.~{Siarohin}, E.~{Sangineto}, S.~R. {Bulò}, N.~{Sebe}, E.~{Ricci},
  Unsupervised domain adaptation using feature-whitening and consensus loss,
  in: Proc. IEEE Conf. Comput. Vis. Pattern Recognit., 2019, pp. 9463--9472.

\bibitem{dan}
M.~{Long}, Y.~{Cao}, Z.~{Cao}, J.~{Wang}, M.~I. {Jordan}, Transferable
  representation learning with deep adaptation networks, IEEE Trans. Pattern
  Anal. Mach. Intell. 41 (2019) 3071--3085.

\bibitem{jan}
M.~Long, H.~Zhu, J.~Wang, M.~I. Jordan, Deep transfer learning with joint
  adaptation networks, in: Proc. Int. Conf. Mach. Learn., 2017, pp. 2208--2217.

\bibitem{tpn}
Y.~{Pan}, T.~{Yao}, Y.~{Li}, Y.~{Wang}, C.~{Ngo}, T.~{Mei}, Transferrable
  prototypical networks for unsupervised domain adaptation, in: Proc. IEEE
  Conf. Comput. Vis. Pattern Recognit., 2019, pp. 2234--2242.

\bibitem{gpda}
M.~{Kim}, P.~{Sahu}, B.~{Gholami}, V.~{Pavlovic}, Unsupervised visual domain
  adaptation: A deep max-margin gaussian process approach, in: Proc. IEEE Conf.
  Comput. Vis. Pattern Recognit., 2019, pp. 4375--4385.

\bibitem{associativeDA}
P.~{Haeusser}, T.~{Frerix}, A.~{Mordvintsev}, D.~{Cremers}, Associative domain
  adaptation, in: Proc. IEEE Int. Conf. Comput. Vis., 2017, pp. 2784--2792.

\bibitem{adr}
K.~Saito, Y.~Ushiku, T.~Harada, K.~Saenko, Adversarial dropout regularization,
  in: Proc. Int. Conf. on Learn. Rep., 2018.

\bibitem{mcd}
K.~{Saito}, K.~{Watanabe}, Y.~{Ushiku}, T.~{Harada}, Maximum classifier
  discrepancy for unsupervised domain adaptation, in: Proc. IEEE Conf. Comput.
  Vis. Pattern Recognit., 2018, pp. 3723--3732.

\bibitem{swd}
C.~{Lee}, T.~{Batra}, M.~H. {Baig}, D.~{Ulbricht}, Sliced wasserstein
  discrepancy for unsupervised domain adaptation, in: Proc. IEEE Conf. Comput.
  Vis. Pattern Recognit., 2019, pp. 10277--10287.

\bibitem{pfan}
C.~{Chen}, W.~{Xie}, W.~{Huang}, Y.~{Rong}, X.~{Ding}, Y.~{Huang}, T.~{Xu},
  J.~{Huang}, Progressive feature alignment for unsupervised domain adaptation,
  in: Proc. IEEE Conf. Comput. Vis. Pattern Recognit., 2019, pp. 627--636.

\bibitem{dirt_t}
R.~Shu, H.~Bui, H.~Narui, S.~Ermon, A {DIRT}-t approach to unsupervised domain
  adaptation, in: Proc. Int. Conf. on Learn. Rep., 2018.

\bibitem{kernel_two_sample_test}
A.~Gretton, K.~M. Borgwardt, M.~J. Rasch, B.~Sch\"{o}lkopf, A.~Smola, A kernel
  two-sample test, Journ. of Mach. Learn. Res. 13 (2012) 723–773.

\bibitem{MMD_D}
F.~Liu, W.~Xu, J.~Lu, G.~Zhang, A.~Gretton, D.~Sutherland, Learning deep
  kernels for non-parametric two-sample tests, in: Proc. Int. Conf. Mach.
  Learn., 2020.

\bibitem{gaacn}
W.~Chen, H.~Hu, Generative attention adversarial classification network for
  unsupervised domain adaptation, Pattern Recognition 107 (2020) 107440.

\bibitem{ctsn}
L.~Zuo, M.~Jing, J.~Li, L.~Zhu, K.~Lu, Y.~Yang, Challenging tough samples in
  unsupervised domain adaptation, Pattern Recognition (2020) 107540.

\bibitem{simultaneous_transfer}
E.~Tzeng, J.~Hoffman, T.~Darrell, K.~Saenko, Simultaneous deep transfer across
  domains and tasks, in: Proc. IEEE Int. Conf. Comput. Vis., 2015, pp.
  4068--4076.

\bibitem{atda}
K.~Saito, Y.~Ushiku, T.~Harada, Asymmetric tri-training for unsupervised domain
  adaptation, in: Proc. Int. Conf. Mach. Learn., Vol.~70, 2017, pp. 2988--2997.

\bibitem{ican}
W.~{Zhang}, W.~{Ouyang}, W.~{Li}, D.~{Xu}, Collaborative and adversarial
  network for unsupervised domain adaptation, in: Proc. IEEE Conf. Comput. Vis.
  Pattern Recognit., 2018, pp. 3801--3809.

\bibitem{dsbn}
W.~{Chang}, T.~{You}, S.~{Seo}, S.~{Kwak}, B.~{Han}, Domain-specific batch
  normalization for unsupervised domain adaptation, in: Proc. IEEE Conf.
  Comput. Vis. Pattern Recognit., 2019, pp. 7346--7354.

\bibitem{pseudo_label}
D.-H. Lee, Pseudo-label : The simple and efficient semi-supervised learning
  method for deep neural networks, in: Workshop of Proc. Int. Conf. Mach.
  Learn., 2013.

\bibitem{da_for_or}
R.~Gopalan, R.~Li, R.~Chellappa, Domain adaptation for object recognition: An
  unsupervised approach, in: Proc. IEEE Int. Conf. Comput. Vis., 2011.

\bibitem{dlid}
S.~Chopra, S.~Balakrishnan, Dlid: Deep learning for domain adaptation by
  interpolating between domains, in: Workshop of Proc. Int. Conf. Mach. Learn.,
  2013.

\bibitem{dm_ada}
M.~Xu, J.~Zhang, B.~Ni, T.~Li, C.~Wang, Q.~Tian, W.~Zhang, Adversarial domain
  adaptation with domain mixup, in: Association for the Advancement of
  Artificial Intelligence, 2020, pp. 6502--6509.

\bibitem{office31}
K.~Saenko, B.~Kulis, M.~Fritz, T.~Darrell, Adapting visual category models to
  new domains, in: Proc. Eur. Conf. Comput. Vis., 2010, pp. 213--226.

\bibitem{cluster_review}
A.~K. Jain, M.~N. Murty, P.~J. Flynn, Data clustering: A review, ACM Computing
  Surveys 31 (1999) 264--323.

\bibitem{clan}
Y.~{Luo}, L.~{Zheng}, T.~{Guan}, J.~{Yu}, Y.~{Yang}, Taking a closer look at
  domain shift: Category-level adversaries for semantics consistent domain
  adaptation, in: Proc. IEEE Conf. Comput. Vis. Pattern Recognit., 2019, pp.
  2502--2511.

\bibitem{officehome}
H.~Venkateswara, J.~Eusebio, S.~Chakraborty, S.~Panchanathan, Deep hashing
  network for unsupervised domain adaptation, in: Proc. IEEE Conf. Comput. Vis.
  Pattern Recognit., 2017, pp. 5385--5394.

\bibitem{visda2017}
The visda-2017 dataset is available at \url{http://ai.bu.edu/visda-2017/}.

\bibitem{svhn}
Y.~Netzer, T.~Wang, A.~Coates, A.~Bissacco, B.~Wu, A.~Y. Ng, Reading digits in
  natural images with unsupervised feature learning, in: Workshop of Proc.
  Neur. Info. Proc. Sys., 2011.

\bibitem{usps}
J.~J. Hull, A database for handwritten text recognition research, IEEE Trans.
  Pattern Anal. Mach. Intell. 16 (1994) 550–554.

\bibitem{em}
Y.~Grandvalet, Y.~Bengio, Semi-supervised learning by entropy minimization, in:
  Proc. Neur. Info. Proc. Sys., 2004, pp. 529--536.

\bibitem{t_sne}
L.~van~der Maaten, G.~Hinton, Visualizing data using t-sne, Journ. of Mach.
  Learn. Res. 9 (2008) 2579–2605.

\bibitem{consist_enforce}
M.~Sajjadi, M.~Javanmardi, T.~Tasdizen, Regularization with stochastic
  transformations and perturbations for deep semi-supervised learning, in:
  Proc. Neur. Info. Proc. Sys., 2016, pp. 1163--1171.

\bibitem{larger_norm}
R.~{Xu}, G.~{Li}, J.~{Yang}, L.~{Lin}, Larger norm more transferable: An
  adaptive feature norm approach for unsupervised domain adaptation, in: Proc.
  IEEE Int. Conf. Comput. Vis., 2019, pp. 1426--1435.

\bibitem{cat}
Z.~{Deng}, Y.~{Luo}, J.~{Zhu}, Cluster alignment with a teacher for
  unsupervised domain adaptation, in: Proc. IEEE Int. Conf. Comput. Vis., 2019,
  pp. 9943--9952.

\bibitem{selfensembling}
G.~French, M.~Mackiewicz, M.~Fisher, Self-ensembling for visual domain
  adaptation, in: Proc. Int. Conf. on Learn. Rep., 2018.

\end{thebibliography}
    \end{small}

	\noindent\textbf{Hui Tang} received the B.Eng. degree in School of Electronic and Information Engineering from South China University of Technology, China, in 2018. She is currently pursuing the Ph.D. degree in School of Electronic and Information Engineering from South China University of Technology. Her research interests are in computer vision and pattern recognition.
	
	\noindent\textbf{Kui Jia} received the B.Eng. degree in marine engineering from Northwestern Polytechnical University, China, in 2001, the M.Eng. degree in electrical and computer engineering from National University of Singapore in 2003, and the Ph.D. degree in computer science from Queen Mary University of London, U.K., in 2007. He is currently a professor in School of Electronic and Information Engineering from South China University of Technology. His research interests are in computer vision, machine learning, and image processing.
	
\end{document}